\documentclass[lettersize,journal]{IEEEtran}
\usepackage{amsmath,amsfonts}
\usepackage[ruled,boxed,linesnumbered]{algorithm2e}
\usepackage{array}
\usepackage[caption=false,font=normalsize,labelfont=sf,textfont=sf]{subfig}
\usepackage{textcomp}
\usepackage{stfloats}
\usepackage{url}
\usepackage{verbatim}
\usepackage{graphicx}
\usepackage{cite}
\usepackage[colorlinks,urlcolor=red]{hyperref}

\usepackage[utf8]{inputenc}
\usepackage[small]{caption}
\usepackage{graphicx}
\usepackage{amsmath}
\usepackage{amsthm}
\usepackage{booktabs}
\usepackage{wasysym}
\usepackage{amssymb}
\usepackage{amsfonts}       
\usepackage{multirow}
\usepackage{multicol}
\usepackage{nicefrac}       
\usepackage{microtype}      
\usepackage{graphicx}
\usepackage{comment}
\usepackage{booktabs}
\usepackage{color}
\usepackage{xcolor}
\usepackage{}

\graphicspath{{figs/}}

\def\mb{\mathbf}
\def\mbb{\mathbb}
\def\mc{\mathcal}

\def\etal{{\em et al.\/}\, }

\def\ie{\textit{i.e.}}

\usepackage{multirow}

\usepackage{colortbl}

\definecolor{mygray}{gray}{.92}

\definecolor{mypink}{rgb}{.99,.91,.95}

\definecolor{mycyan}{cmyk}{.3,0,0,0}

\begin{document}

\title{Multi-Label Continual Learning \\ using Augmented Graph Convolutional Network}

\author{Kaile Du$^*$, Fan Lyu$^*$,~\IEEEmembership{Student Member,~IEEE,}
Linyan Li,
Fuyuan Hu$^\dag$,~\IEEEmembership{Member,~IEEE,}\\
Wei Feng,~\IEEEmembership{Member,~IEEE,}
Fenglei Xu, Xuefeng Xi, Hanjing Cheng
\thanks{\IEEEcompsocthanksitem K. Du, F. Hu, F. Xu, X. Xi and H. Cheng are with the Suzhou University of Science and Technology, Suzhou 215009, China.\protect\\
  E-mail: \{kailedu@post, fuyuanhu@mail, xufl@mail, xfxi@mail, chj@mail\}.usts.edu.cn
  \IEEEcompsocthanksitem F. Lyu and W. Feng are with the College of Intelligence and Computing, Tianjin University, Tianjin 300350, China.\protect\\
  E-mail: \{fanlyu, wfeng\}@tju.edu.cn
  \IEEEcompsocthanksitem L. Li is with Suzhou Institute of Trade \& Commerce, Suzhou 215009, China.\protect\\
  E-mail: lilinyan@szjm.edu.cn
  \IEEEcompsocthanksitem $^*$Contributes equally.
  \IEEEcompsocthanksitem $^\dag$Corresponding author: Fuyuan Hu.}
}

\markboth{Journal of \LaTeX\ Class Files,~Vol.~14, No.~8, August~2021}%
{Shell \MakeLowercase{\textit{et al.}}: A Sample Article Using IEEEtran.cls for IEEE Journals}


\maketitle

\begin{abstract}

  Multi-Label Continual Learning (MLCL) builds a class-incremental framework in a sequential multi-label image recognition data stream.
  The critical challenges of MLCL are the construction of label relationships on \textit{past-missing and future-missing partial labels} of training data and the \textit{catastrophic forgetting} on old classes, resulting in poor generalization.	
  To solve the problems, the study proposes an Augmented Graph Convolutional Network (AGCN++) that can construct the cross-task label relationships in MLCL and sustain catastrophic forgetting.
  First, we build an Augmented Correlation Matrix (ACM) across all seen classes, where the intra-task relationships derive from the hard label statistics. In contrast, the inter-task relationships leverage hard and soft labels from data and a constructed expert network.
	{Then, we propose a novel partial label encoder (PLE) for MLCL, which can extract dynamic class representation for each partial label image as graph nodes and help generate soft labels to create a more convincing ACM and suppress forgetting.}
	Last, to suppress the forgetting of label dependencies across old tasks, we propose a relationship-preserving constrainter to construct label relationships.
 The inter-class topology can be augmented automatically, which also yields effective class representations.
	The proposed method is evaluated using two multi-label image benchmarks. The experimental results show that the proposed way is effective for MLCL image recognition and can build convincing correlations across tasks even if the labels of previous tasks are missing.
\end{abstract}


\maketitle

\section{Introduction}

%
%
%
%

%
\IEEEPARstart{M}ACHINE learning approaches have been reported to exhibit human-level performance on some tasks, such as Atari games~\cite{silver2018general}  or object recognition~\cite{russakovsky2015imagenet}.
However, they always assume that no novel knowledge will be input into models, which is impractical in the real world.
To meet the scenario, continual learning develops intelligent systems that can continuously learn new tasks from sequential datasets while preserving learned knowledge of old tasks~\cite{chen2018lifelong}. 
Recently, class-incremental continual learning~\cite{rebuffi2017icarl} builds an adaptively evolvable classifier for the seen classes at any time, where the learner has no access to the task-ID at inference time~\cite{van2019three} just like the real-life applications. 
{Compared to traditional continual learning, a class-incremental model has to distinguish between all seen classes from all tasks. Therefore is more challenging. }
For privacy and storage reasons, the training data for old tasks is unavailable when new tasks arrive.
As the model incrementally learns new knowledge, old knowledge is overwritten and gets a drop in performance, known as catastrophic forgetting~\cite{kirkpatrick2017overcoming}. 
Thus, the major challenge of MLCL is to learn new tasks without catastrophically forgetting previous tasks over time.

\begin{figure}[t]
	\centering
	\includegraphics[width=\linewidth]{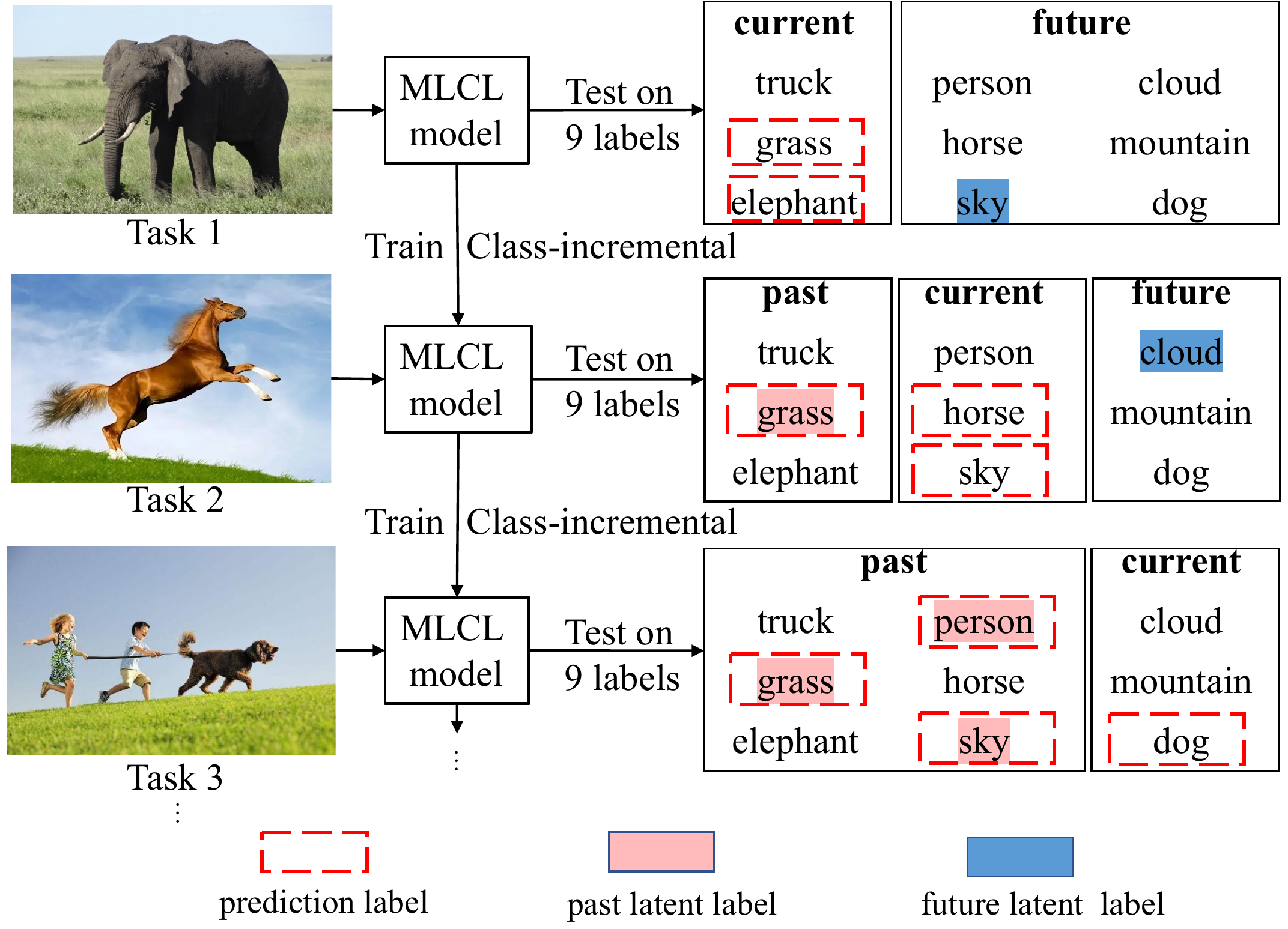}
	\caption{
          The inference process of MLCL.
          An MLCL model can recognize more labels by learning incremental classes given multi-label images. With learning three classes at each task, the MLCL model can recognize these labels continuously.
	}
	\label{fig:lml}
\end{figure}


Due to many researchers' efforts, many methods for class-incremental continual learning have been proposed. 
The rehearsal-based methods~\cite{mai2021supervised,bang2021rainbow,lyu2020multi,kim2020imbalanced,sun2022exploring} stores samples from raw datasets or generates pseudo-samples with a generative model~\cite{9161395,ye2021lifelong}, these samples are replayed while learning a new task to prevent forgetting.
The regularization-based methods~\cite{9552512,9757872,kirkpatrick2017overcoming,li2017learning,zhou2022few} have an additional regularization term introduced in the loss function, consolidating previous knowledge when learning on new tasks.
And the parameter isolation methods~\cite{mallya2018packnet,mallya2018piggyback,aljundi2017expert} dedicates different model parameters to each task to alleviate any possible forgetting. Moreover, some recent transformer-based methods~\cite{zhou2022learning,douillard2022dytox,wang2022dualprompt,wang2022sprompt} have also achieved good performance.

However, most existing methods for class-incremental continual learning only consider the input images are {single-labelled}, and we call them Single-Label Continual Learning (SLCL).
SLCL is limited in practical applications such as movie categorization and scene classification, where \emph{multi-label} data is widely used.
As shown in Fig.~\ref{fig:lml}, an image contains multiple labels, including ``sky'', ``grass'', ``person'' and ``dog'', etc., which shows the multi-label space is much larger than the single-label space via label combination.
Because of the co-occurrence of multiple labels, the label space of the multi-label dataset is much larger than that of the single-label one.
In the recent, using a neural network to tackle Multi-Label (ML) image classification problems has achieved impressive results~\cite{zhu2017learning,li2017improving}, which consider constructing label relationships to improve the classification by using recurrent neural network~\cite{wang2016cnn,lyu2019attend} and graph convolutional network~\cite{chen2019multi,chen2020knowledge,chen2021learning}.

This paper puts the multi-task classification into an incremental scenario, \ie, class-incremental classification, and studies how to sequentially learn new classes for Multi-label Continual Learning (MLCL). 
The inference process of MLCL is in Fig.~\ref{fig:lml}. 
Given testing images, the model can incrementally recognize multiple labels as new classes are learned continuously.
Because of the unavailability of data with future and past classes in continual learning, the \emph{partial label problem} poses a significant challenge in building multi-label relationships and reducing catastrophic forgetting in MLCL than SLCL.
The partial label problem means that each task in MLCL cannot be trained independently since the label spaces for different tasks are overlapped. 
For example, as shown in Fig.~\ref{fig:lml}, the label ``sky'' is present in all three tasks. It is one of the overlapped labels for three tasks. 
For task $1$, ``sky'' is the future latent label, and for task $3$, "sky" is the past latent label. If the past latent label is not annotated in the current training, the past-missing partial label problem will occur, and similarly, the future-missing partial label problem will occur.
An MLCL model should incrementally recognize multiple labels as new classes are learned continuously.

Practically, we solve the MLCL problems in two real-world labelling scenarios, \ie, the current training dataset has past and current labels (Continuous Labelling MLCL, \textit{CL-MLCL}) or only current labels (Independent Labelling MLCL, \textit{IL-MLCL}). 
The IL-MLCL has past-missing and future-missing partial label problems, while CL-MLCL has only the future-missing partial label problem.
As for both scenarios, the partial label problem poses a significant challenge in building multi-label relationships and keeping them from catastrophic forgetting.
It is crucial to study a feasible solution to solve the partial label problem in MLCL.
\textit{This motivates us to design a unified MLCL solution to the sequential multi-label classification problem by considering the label relationships across tasks in both IL-MLCL and CL-MLCL scenarios.}

This paper is an extension of our previous work, Augmented Graph Convolutional Network (AGCN)~\cite{9859622}, and we complete the real-world scenario of MLCL (IL-MLCL and CL-MLCL) and propose an improved version, AGCN++.
Our AGCN++ has three major parts.
First, to relate partial labels across tasks, we propose to construct an Augmented Correlation Matrix (ACM) sequentially in MLCL. We design a unified ACM constructor.
For CL-MLCL, ACM is updated by the hard label statistics from new training data at each task.
For IL-MLCL, an auto-updated expert network is designed to generate predictions of the old tasks. These predictions are used as soft labels to represent the old classes in constructing ACM.
{Second, due to partial label problems, effective class representation is difficult to build. In our early conference work, the AGCN model utilized pre-given semantic information (\ie~word embedding) as class representation. 
The fixed class representation will lead to the accumulation of errors in constructing label relationships due to partial label problems. Then, this will skew predictions and lead to more serious forgetting. 
So in this work, AGCN++ utilizes a partial label encoder (PLE) to decompose each partial image's feature into dynamic class representations. These class-specific representations will vary from image to image and are input as graph nodes into AGCN++. Moreover, unlike AGCN, which directly adds graph nodes manually, PLE can automatically generate graph nodes for each partial label image. Utilizing PLE to get the graph nodes can also reduce the impact of the low quality of word embeddings. So AGCN++ can generate a more convincing ACM and suppress forgetting.}
Third, we propose to encode the dynamically constructed ACM and graph nodes. The AGCN++ model correlates the label spaces of both the old and new tasks in a convolutional architecture and mines the latent correlation for every two classes. This information will be combined with the visual features for prediction.
Moreover, to further mitigate the forgetting, a distillation loss function and a relationship-preserving graph loss function are designed for class-level forgetting and relationship-level forgetting, respectively.

In this paper, we construct two multi-label image classification datasets, Split-COCO and Split-WIDE, based on widely-used multi-label datasets MS-COCO and NUS-WIDE. 
The results on Split-COCO and Split-WIDE show that the proposed AGCN++ effectively reduces catastrophic forgetting for MLCL image recognition and can build convincing correlation across tasks whenever the labels of previous tasks are missing (IL-MLCL) or not (CL-MLCL).
Moreover, our methods can effectively reduce catastrophic forgetting in two scenarios.

This paper extends our AGCN~\cite{9859622} with the following new contents:
\begin{enumerate} 
\item We complete the real-world scenario of MLCL from IL-MLCL to CL-MLCL scenarios, and a unified AGCN++ model is redesigned to capture label dependencies to improve multi-label recognition in the data stream; 
\item We propose a novel partial label encoder (PLE) to decompose the global image features into dynamic graph nodes for each partial label image, which reduces the accumulation of errors in the construction of label relationships and suppresses forgetting; %
\item We propose a unified ACM constructor. The ACM is dynamically constructed using soft or hard labels to build label relationships across sequential tasks of MLCL to solve the partial label problem for IL-MLCL and CL-MLCL. The distillation loss and relationship-preserving loss readjust to IL-MLCL and CL-MLCL to mitigate the class- and relationship-level catastrophic forgetting;
\item More experimental results are provided, including {extensive comparisons on two different scenarios settings and more ablation studies, etc.} More ablation studies and new SOTA MLCL results are provided.
\end{enumerate}




\section{Related Work}



\subsection{Class-incremental continual learning}

Class-incremental continual learning~\cite{rebuffi2017icarl} builds a classifier that learns a sequence of new tasks corresponding to different classes.
The state-of-art methods for class-incremental continual learning can be categorized into three main branches to solve the catastrophic forgetting problem.

First, the regularization-based methods~\cite{li2017learning,kirkpatrick2017overcoming,zhou2022few,9552512,9757872}, this line of work introduces additional regularization terms in the loss function to consolidate previous knowledge when learning new tasks.
These methods are based on regularizing the parameters corresponding to the old tasks, penalizing the feature drift on the old tasks and avoiding storing raw inputs.
Kirkpatrick \etal\cite{kirkpatrick2017overcoming} limits changes to parameters based on their significance to the previous tasks using Fisher information;
LwF~\cite{li2017learning} is a data-focused method, and it leverages the knowledge distillation combined with a standard cross-entropy loss to mitigate forgetting and transfer knowledge by storing the previous parameters. 
Thuseethan \etal\cite{9552512} propose an indicator loss, which is associated with the distillation mechanism that preserves the existing upcoming emotion knowledge.
Yang \etal\cite{9757872} introduce an attentive feature distillation approach to mitigate catastrophic forgetting while accounting for semantic spatial- and channel-level dependencies. The regularization-based procedures can protect privacy better because they do not collect samples from the original dataset.

Second, the rehearsal-based methods~\cite{ye2021lifelong,de2021continual,bang2021rainbow,lyu2020multi,rolnick2019experience,rebuffi2017icarl,chaudhry2018efficient,silver2020generating,shin2017continual}, which sample a limited subset of data from the previous tasks or a generative model as the memory.
The stored memory is replayed while learning new tasks to mitigate forgetting.
In ER~\cite{rolnick2019experience}, this memory is retrained as the extended training dataset during the current training;
RM~\cite{bang2021rainbow} is a replay method for the blurry setting; 
iCaRL~\cite{rebuffi2017icarl} selects and stores samples closest to the feature mean of each class for replaying;
AGEM~\cite{chaudhry2018efficient} resets the training gradient by combining the gradient on the memory and training data; 
Ye \etal\cite{ye2021lifelong} propose a Teacher-Student network framework. The Teacher module would remind the Student about the information learnt in the past.

Third, the parameter isolation based methods~\cite{aljundi2017expert,mallya2018packnet,mallya2018piggyback,serra2018overcoming}, which generate task-specific parameter expansion or sub-branch. When no limits apply to the size of networks, Expert Gate~\cite{aljundi2017expert} grows new branches for new tasks by dedicating a model copy to each task. PackNet~\cite{mallya2018packnet} iteratively assigns parameter subsets to consecutive tasks by constituting binary masks.

Though the existing methods have achieved successes in SLCL, they are hardly used in MLCL directly.
The overlook of the partial label problem means the inevitability of more serious forgetting in MLCL, let alone the construction of multi-label relationships and reducing the forgetting of the relationships.

\subsection{Multi-label image classification}

Compared with the traditional single-label classification problem, multi-label classification is more practical in the real world.
Earlier multi-label learning methods~\cite{yang2016exploit} prefer to build the model with the help of extra-label localisation information, which is assumed to include all possible foreground objects. And they aggregate the features from proposals to incorporate local information for multi-label prediction.
However, extra localisation information is costly, preventing the models from applying to end-to-end training approaches.
More recent advances are mainly by constructing the label relationships.
Some works~\cite{wang2016cnn,lyu2019attend} use the recurrent neural network (RNN) for multi-label recognition under a restrictive assumption that the label relationships are in order, which limits the complex relationships in label space. 
Furthermore, some works \cite{chen2019multi,chen2020knowledge,chen2021learning} build label relationships using graph structure and use graph convolutional network (GCN) to enhance the representation.
The standard limit of these methods is that they can only construct the intra-task correlation matrix using the training data from the current task and fail to capture the inter-task label dependencies in a continual data stream.
They rely on prior knowledge to construct the correlation matrix, which is the key of GCN that aims to gain the label dependencies. These methods utilize the information of the whole training dataset to capture the co-occurrence patterns of objects in an offline way.
 Some recent methods focus on the partial label problem~\cite{gong2021top,9343691,9529072} for offline multi-label learning. Compared with this offline way, we construct the correlations and use soft label statistics to solve the partial label problem for MLCL. 
Kim \etal\cite{kim2020imbalanced} propose to extend the ER~\cite{rolnick2019experience} algorithm using an improved reservoir sampling strategy to study the imbalanced problem on multi-label datasets.
However, the label dependencies are ignored in this work \cite{kim2020imbalanced}. 
In contrast, we propose to model label relationships sequentially in MLCL and consider mitigating the relationship-level forgetting in MLCL.


\section{Multi-Label Continual Learning}

\begin{figure*}[t]
	\centering
	\includegraphics[width=\linewidth]{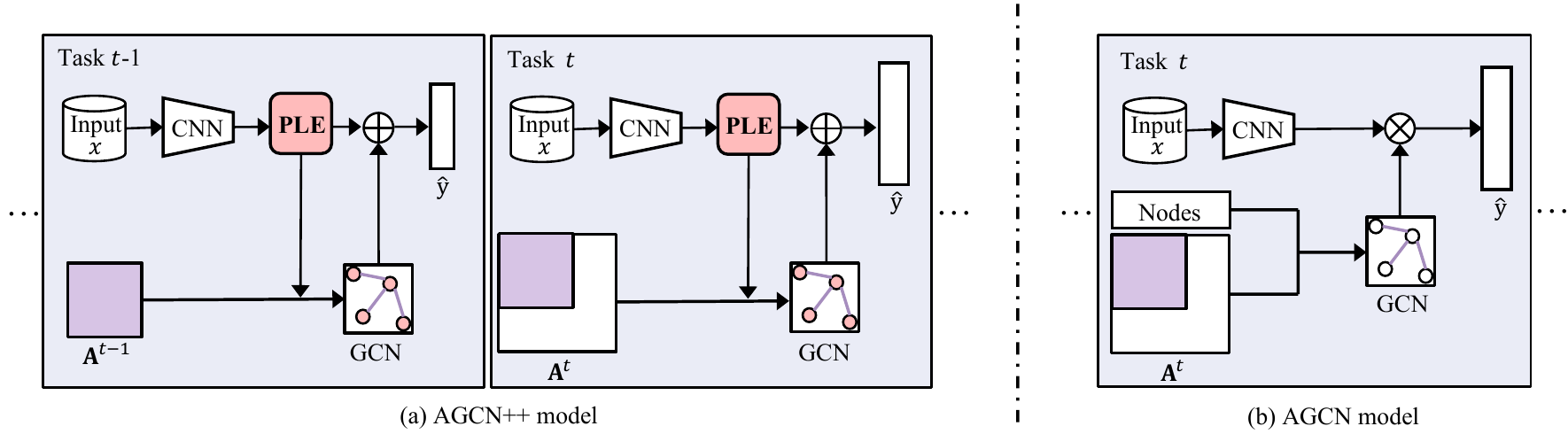}
	\caption{
          The overall framework of AGCN++ and AGCN.
          (a) AGCN++ model.  The model is mainly composed of three points: PLE, ACM, and stacked GCN encoder. PLE decomposes the image feature extracted by the CNN into a group of class-specific representations. The stacked GCN encodes these representations and ACM into graph features. $\hat{y}$ denotes the class-incremental prediction scores.
          (b) Compared with AGCN++, AGCN directly uses word embedding as graph nodes. Graph nodes and ACM are input into GCN. The graph feature and image feature do the matrix multiplication to get the prediction.
          }
	\label{fig:framework}
\end{figure*}

\begin{figure}[t]
	\centering
	\includegraphics[width=0.78\linewidth]{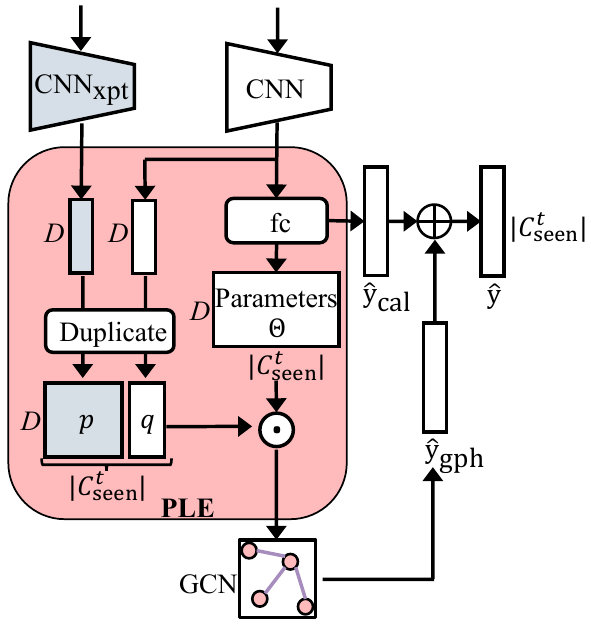}
	\caption{Partial label encoder (PLE) in task $t$.}
	\label{fig:PLE}
\end{figure}




\subsection{Definition of MLCL}
Given $T$ tasks with respect to training datasets $\{\mc{D}^{1}_\text{trn},\cdots,\mc{D}^{T}_\text{trn}\}$ and test datasets $\{\mc{D}^{1}_\text{tst},\cdots,\mc{D}^{T}_\text{tst}\}$, the total class numbers increase gradually with the sequential tasks in MLCL and the model is constantly learning new knowledge.
A continual learning system trains on the training sets from $\mc{D}^{1}_\text{trn}$ to $\mc{D}^{T}_\text{trn}$ sequentially and evaluate on all seen test sets at any time.
For the $t$-th new task, the new and task-specific classes are to be trained, namely $\mc{C}^t$. 
\textit{MLCL aims at learning a multi-label classifier to discriminate the increasing number of classes in the continual learning process.}
We denote $\mc{C}_\text{seen}^t=\bigcup_{n=1}^{t}\mc{C}^n$ as seen classes at task $t$, which  contains old class set $\mc{C}_\text{seen}^{t-1}$ and new class set $\mc{C}^t$, that is, $\mc{C}_\text{seen}^t=\mc{C}_\text{seen}^{t-1}\cup\mc{C}^t$, and $\mc{C}_\text{seen}^{t-1}\cap\mc{C}^t=\varnothing$.

\subsection{MLCL scenarios}
In this section, considering academic and practical requirements, we introduce the two scenarios in MLCL. 
In one scenario, we adopt strict continual learning and cannot obtain the old class like most single-Label continual learning methods~\cite{lyu2020multi,kim2020imbalanced,9161395,ye2021lifelong}. In the other scenario, we consider the real-world setting.
IL-MLCL setting has hard task boundaries, so the old classes are unavailable. Conversely, similar to the settings in~\cite{bang2021rainbow}  and~\cite{aljundi2019gradient}, CL-MLCL setup makes the task boundaries faint. It is closer to the real world, where new classes do not show up exclusively. The difference between the two scenarios is the training label space for old classes.

\subsubsection{Continuous labelling (CL-MLCL)}

CL-MLCL is a more realistic scenario where the data distribution shifts gradually without hard task boundaries. The annotator of CL-MLCL needs to label all seen classes. 
The class numbers of training data increase gradually with the sequential tasks, \ie, $\mc{C}_\text{seen}^t$ for training data of task $t$. As shown in Table~\ref{tab:scenarios}, the label space $\mc{Y}\subseteq\mc{C}_\text{seen}^t$.
The past latent label is annotated, so the old and new labels coexist for a current sample in CL-MLCL.
This scenario is labor-costly, especially when the class number is large. {Because the past latent label is annotated, only the \emph{future-missing partial label} problem will occur in CL-MLCL, and no past-missing partial label problem will occur.}



\subsubsection{Independent labelling (IL-MLCL)}
In this scenario, the annotator only labels the new classes in $\mc{C}^t$ for training data in task $t$, as shown in Table~\ref{tab:scenarios}.
This means the training label space is independently labelled with sequential class-incremental tasks.
The old and new labels do not overlap in new task samples in IL-MLCL.
The training label space $\mc{Y}$ of IL-MLCL at task $t$ is right the task-specific label (new labels) set $\mc{C}^t$.
IL-MLCL can reduce the labelling cost, but due to the lack of old labels in the IL-MLCL label space, the past latent label is not annotated, so a \emph{past-missing partial label} problem will be caused together with \emph{future-missing partial label}.



\subsubsection{Test phase and the goal}
During the test phase, the ground truth for each data point contains all the old classes $\mc{C}_\text{seen}^{t-1}$ and task-specific classes $\mc{C}^t$ for both CL-MLCL and IL-MLCL.
That is, as shown in Table~\ref{tab:scenarios}, the label space in the test phase is the all seen classes $\mc{C}_\text{seen}^t$.
This paper aims to propose a unified approach to solve the MLCL problem in both IL-MLCL and CL-MLCL scenarios.

\begin{table}[t]
  \centering
  \caption{
    Training and testing label sets of task $t$ in two scenarios, CL-MLCL and IL-MLCL.
  }
  \resizebox{.8\linewidth}{!}{
    \begin{tabular}{c|c|c}
      \toprule
      & \textbf{CL-MLCL} & \textbf{IL-MLCL} \\ 
      \midrule
      Trian & $\mc{C}_\text{seen}^t=\mc{C}_\text{seen}^{t-1}\cup\mc{C}^t$    & $\mc{C}^t$  \\ 
      Test  & $\mc{C}_\text{seen}^t=\mc{C}_\text{seen}^{t-1}\cup\mc{C}^t$    & $\mc{C}_\text{seen}^t=\mc{C}_\text{seen}^{t-1}\cup\mc{C}^t$   \\ 
      \bottomrule
    \end{tabular}}
  \label{tab:scenarios}
\end{table}

\section{Methodology}

\subsection{Overview of the proposed method}
In multi-label learning, label relationships are verified effective to improve the recognition~\cite{chen2019multi,chen2020knowledge,chen2021learning}.
However, it is challenging to construct convincing label relationships in MLCL image recognition because of the \textit{partial label} problem. 
The partial label problem results in difficulty in constructing the inter-task label relationships.
Moreover, forgetting happens not only at the class level but also at the relationship level, which may damage performance. 

For effective multi-label recognition, we propose an AGCN++ to construct and update the intra- and inter-task label relationships during the training process.
As shown in Fig.~\ref{fig:framework} (a), AGCN++ model is mainly composed of three parts:
1) {\textbf{Partial label encoder (PLE)} decomposes the image feature extracted by the CNN into a group of class-specific representations, these representations are used as graph nodes to feed the GCN model.}
2) \textbf{Augmented Correlation Matrix (ACM)}  provides the label relationships among all seen classes $\mc{C}^t_{\text{seen}}$ and is augmented to capture the intra- and inter-task label dependencies.
3) Graph Convolutional Network encodes \textbf{ACM} and graph nodes \textbf{$\mb{H}^t$} into label representations $\mb{\hat{y}}_\text{gph}$ for label relationships.
{We construct auto-updated expert networks consisting of $\text{CNN}_\text{xpt}$ and $\text{GCN}_\text{xpt}$. After each task has been trained, the model is saved as the expert model to provide soft labels $\mb{\hat{z}}$}.

As shown in Fig.~\ref{fig:framework} (a) and (b), the most significant difference between AGCN and AGCN++ is that AGCN++ can extract graph nodes from the original image through PLE. GCN encodes ACM and graph nodes to get $\mb{\hat{y}}_\text{gph}$. By adding $\mb{\hat{y}}_\text{cal}$ and $\mb{\hat{y}}_\text{gph}$, the soft label generated by the model can better replace the past-missing partial label, more convincing ACM (see Fig.~\ref{fig:ACM_visual}) can be developed for IL-MLCL, and the forgetting of CL-MLCL and IL-MLCL can be reduced through knowledge distillation. These can improve the performance of the model.

\subsection{Partial label encoder}
\label{sec:ple}
Due to the partial label problems in MLCL, effective class representation is difficult to build. In AGCN, the model utilized pre-given word embedding as fixed class representation, which will lead to the accumulation of errors in the construction of label relationships. Then, this will skew predictions and lead to more serious forgetting.

{Inspired by~\cite{zhou2016learning}, we propose the partial label encoder (PLE), which decomposes the global image features for each partial label into dynamic representations continuously as classes increment. These class-specific representations will vary from image and are used as augmented graph nodes. PLE will reduce the accumulation of errors in the construction of label relationships and suppress forgetting. And the resulting graph nodes are automatically augmented as the number of classes increases in MLCL.}

PLE initializes with image features and model parameters and continuously updates graph nodes. 
As shown in Fig~\ref{fig:PLE}, $\text{CNN}(x)\in\mbb{R}^{D}$, $\text{CNN}_\text{xpt}(x)\in\mbb{R}^{D}$, $D$ represents the image feature dimensionality.
We use a fully connected layer fc$(\cdot)$ to achieve two goals.  
One is to get the prediction without adding label dependencies $\mb{\hat{y}}_\text{cal}$
\begin{equation}\label{eq:cal}
  \mb{\hat{y}}_\text{cal} = \text{fc}(\text{CNN}(x))   \in\mbb{R}^{|\mc{C}_\text{seen}^t|}.
\end{equation}
The other is to make the image feature aware of class information by doing Hadamard Product with its parameters. 
\begin{equation}\label{eq:h}
	\mb{H}^{t} = \Theta \odot \text{cat}(p,q) \in\mbb{R}^{|\mc{C}_\text{seen}^t| \times D},
\end{equation}
where $\odot$ is the Hadamard Product. 
$p$ and $q$ are multiple copies of respective image features. 
$p = \text{Duplicate}(\text{CNN}_\text{xpt}(x))\in\mbb{R}^{|\mc{C}_\text{seen}^{t-1}|\times D}$, $q = \text{Duplicate}(\text{CNN}(x))\in\mbb{R}^{|\mc{C}^{t}|\times D}$. {For example, $\text{CNN}(x)\in\mbb{R}^{1\times D}$  is copied $D$ times to get $q\in\mbb{R}^{|\mc{C}^{t}|\times D}$.} $\Theta\in\mbb{R}^{|\mc{C}_\text{seen}^{t}|\times D}$ represents the {class-specific} fc layer parameters. The dimension of $\Theta$ is continuously expanded to accommodate the class-incremental characteristic in continual learning. 
In Eq.~\eqref{eq:h}, $\mb{H}^{t}$ represents the class-aware graph node and automatically augments as the new task progresses. 
We then encode $\mb{H}^{t}$ by Graph Convolutional Network (GCN) to get graph representation $\mb{\hat{y}}_\text{gph}$.
\begin{equation}\label{eq:gph}
  \mb{\hat{y}}_\text{gph} = \text{GCN}(\mb{A}^t, \mb{H}^{t})\in\mbb{R}^{|\mc{C}_\text{seen}^t|},
\end{equation}
where $\mb{A}^t$ denotes the Augmented Correlation Matrix (ACM, see the next section for details). 
GCN is a two-layer stacked graph model similar to ML-GCN~\cite{chen2019multi,chen2021learning}. 
ACM $\mb{A}^t$ and graph node $\mb{H}^{t}$ can be augmented as the class number increments.
With the established ACM, GCN provides dynamic label relationships to CNN for prediction.

Moreover, we introduce the prediction $\mb{\hat{y}}_\text{cal}$ without adding label dependencies, which is combined with $\mb{\hat{y}}_\text{gph}$ as the final multi-label prediction $\mb{\hat{y}}\in\mbb{R}^{|\mc{C}_\text{seen}^{t}|}$ of our model:
\begin{equation}\label{eq:predict}
  \mb{\hat{y}} = \sigma\left( \mb{\hat{y}}_\text{cal} + \mb{\hat{y}}_\text{gph}\right),
\end{equation}
where $\sigma(\cdot)$ represents the Sigmoid function. 

ACM represents the auto-updated dependency among all seen classes in the MLCL image recognition system.
The next section will introduce how to establish and augment ACM in AGCN++.

\subsection{Augmented Correlation Matrix}
\label{sec:acm}

Most existing multi-label learning algorithms~\cite{chen2019multi,chen2020knowledge,chen2021learning} rely on constructing the inferring label correlation matrix $\mb{A}$ by the hard label statistics among the class set $\mc{C}$: $\mb{A}_{ij}=P(\mc{C}_i|\mc{C}_j)|_{i \neq j}$.
The correlation matrix represents a fully-connected graph.
When a new task comes, the graph should be augmented automatically.
\emph{However, in MLCL, the label correlation matrix is hard to infer directly by statistics because of the partial label problem.}

\begin{figure*}[t]
	\centering
	\includegraphics[width=0.88\linewidth]{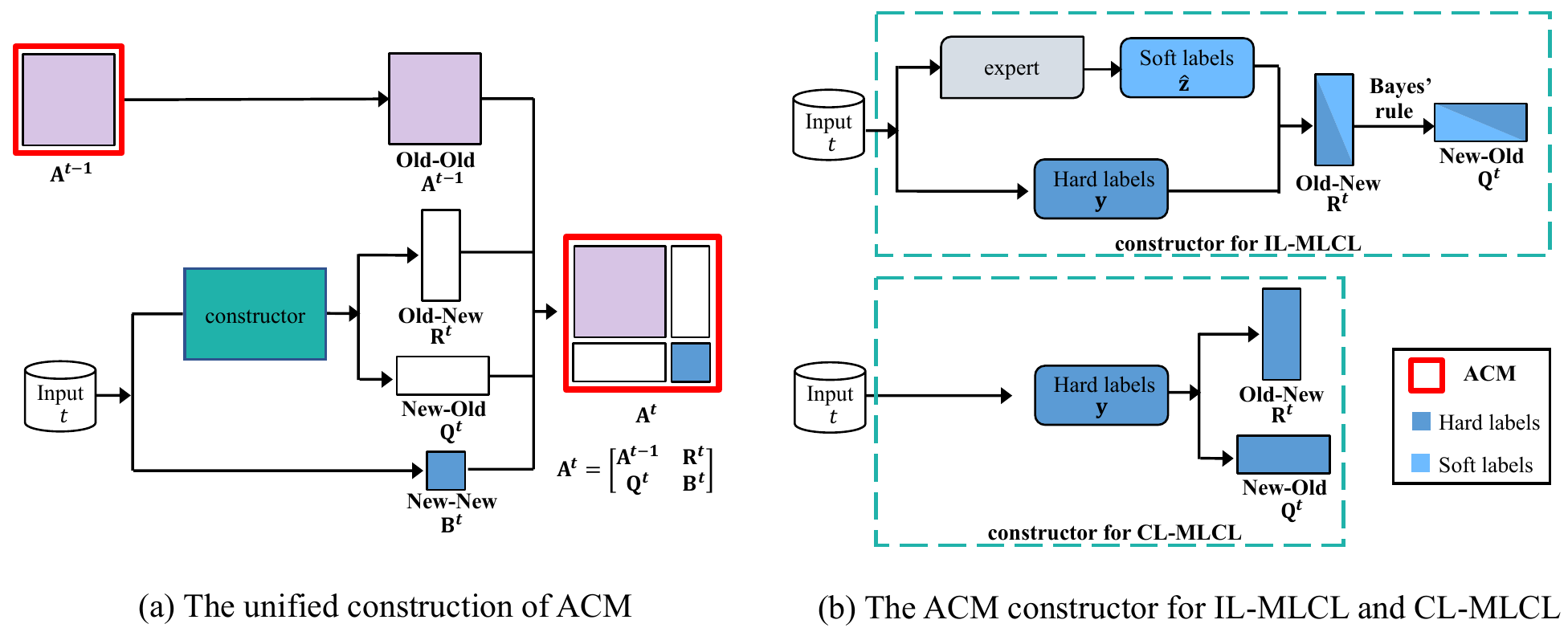}
	\caption{
         (a) The unified construction of ACM $\mb{A}^t$ using Eq.~\eqref{eq:acm}. $\mb{A}^{t-1}$ is the completely-built ACM of task $t-1$ in two scenarios. (b) The ACM constructor for IL-MLCL and CL-MLCL.  For IL-MLCL, $\mb{B}^t$ is constructed using hard labels from a dataset of task $t$ using Eq.~\eqref{eq:hard-hard}; the $\mb{R}^t$ is constructed by hard labels from a dataset of task $t$ and soft labels from the expert using Eq.~\eqref{eq:soft-hard}, the expert is the saved model of task $t-1$; the $\mb{Q}^t$ is constructed by $\mb{R}^t$ based on Bayes' rule using Eq.~\eqref{eq:hard-soft}. For CL-MLCL, the ACM is constructed using hard label statistics in four blocks, including Old-Old, New-New, Old-New and New-Old blocks by Eq.~\eqref{eq:hard-hard}, Eq.~\eqref{eq:soft-hard} and Eq.~\eqref{eq:hard-soft}.}
	\label{fig:ACM_unified}
\end{figure*}

To tackle the problem, as shown in Fig.~\ref{fig:ACM_unified} (b), we introduce an auto-updated expert network inspired by ~\cite{li2017learning} and~\cite{zhou2022few}, which is used to provide missing past labels.
The \textit{soft labels} ${\mb{\hat{z}}}$ is obtained by feeding data of the current task into the expert model, \ie~$\mb{\hat{z}}=\text{expert}(x)$.
Based on the soft labels, as shown in Fig.~\ref{fig:ACM_unified} (a), we construct an Augmented Correlation Matrix (ACM) $\mb{A}^t$ in IL-MLCL and CL-MLCL:
\begin{equation}\label{eq:acm}
	{\mb{A}}^{t}=\begin{bmatrix} {\mb{A}}^{t-1} & \mb{R}^t \\ \mb{Q}^t & \mb{B}^t \end{bmatrix}\iff\begin{bmatrix} \text{Old-Old} & \text{Old-New} \\ \text{New-Old} & \text{New-New} \end{bmatrix},
\end{equation}
in which we take four block matrices including $\mb{A}^{t-1}$ and $\mb{B}^t$, $\mb{R}^t$ and $\mb{Q}^t$ to represent intra- and inter-task label relationships between old and old classes, new and new classes, old and new classes as well as new and old classes respectively.
For the first task, $\mb{A}^1=\mb{B}^1$.
For $t>1$, $\mb{A}^t\in\mbb{R}^{|\mc{C}_\text{seen}^t|\times|\mc{C}_\text{seen}^t|}$.
It is worth noting that the block $\mb{A}^{t-1}$ (Old-Old) can be derived from the old task, so we focus on how to compute the other three blocks in the ACM.

\noindent
\textbf{New-New block} ($\mb{B}^t\in\mbb{R}^{|\mc{C}^t|\times|\mc{C}^t|}$).
As shown in Fig.~\ref{fig:ACM_unified} (a), this block computes the intra-task label relationships among the new classes, and the conditional probability in $\mb{B}^t$ can be calculated using the hard label statistics from the training dataset similar to the common multi-label learning: 
\begin{equation}\label{eq:hard-hard}
	\mb{B}^t_{ij} = P(\mc{C}^t_{i}\in\mc{C}^t|\mc{C}^t_{j}\in\mc{C}^t) = \frac{N_{ij}}{N_{j}},
\end{equation}
where $N_{ij}$ is the number of examples with both class $\mc{C}^t_{i}$ and $\mc{C}^t_{j}$, $N_{j}$ is the number of examples with class $\mc{C}^t_{j}$. 
Due to the data stream, $N_{ij}$ and $N_{j}$ are accumulated and updated at each step of the training process.
This block is shared by both IL- and CL-MLCL because the new class data is always available.



\noindent
\textbf{Old-New block} ($\mb{R}^t\in\mbb{R}^{|\mc{C}_\text{seen}^{t-1}|\times|\mc{C}^t|}$). 
As shown in Fig.~\ref{fig:ACM_unified} (b), for CL-MLCL, this block can be directly obtained by the hard label statistics.
For IL-MLCL, given an image $\mb{x}$, for old classes, ${\hat{z}}_{i}$ (predicted probability) generated by the expert can be considered as the soft label for the $i$-th class. 
Thus, the product ${\hat{z}}_{i}{{{y}}_{j}}$ can be regarded as an alternative of the cooccurrences of ${\mc{C}_\text{seen}^{t-1}}_i$ and $\mc{C}^t_{j}$. 
Thus, $\sum_{\mb{x}}{\hat{z}}_{i}{{{y}}_{j}}$ means the online mini-batch accumulation
\begin{equation}\label{eq:soft-hard}
  \begin{aligned}		
    \mb{R}^t_{ij} 	&= P({\mc{C}_\text{seen}^{t-1}}_i\in{\mc{C}_\text{seen}^{t-1}}|\mc{C}^t_{j}\in\mc{C}^t) \\
    &= 
    \left\{
      \begin{aligned}
        &\frac{N_{ij}}{N_{j}},&\quad &\text{if CL-MLCL},\\
        &\frac{\sum_{\mb{x}}{\hat{z}}_{i}{{{y}}_{j}}}{N_j},&\quad &\text{if IL-MLCL},
      \end{aligned}
    \right.
  \end{aligned}
\end{equation}
where $N_{ij}$ is the accumulated number of examples with both class ${\mc{C}_\text{seen}^{t-1}}_i$ and $\mc{C}^t_{j}$, $N_{j}$ is the accumulated number of examples with class $\mc{C}^t_{j}$.

\noindent
\textbf{New-Old block} ($\mb{Q}^t\in\mbb{R}^{|\mc{C}^t|\times|\mc{C}_\text{seen}^{t-1}|}$). 
As shown in Fig.~\ref{fig:ACM_unified} (b), for CL-MLCL, the inter-task relationship between new and old classes can be computed using hard label statistics.
For IL-MLCL, based on the Bayes' rule, we can obtain this block by
\begin{equation}\label{eq:hard-soft}
  \begin{aligned}
    \mb{Q}^t_{ji} &= P(\mc{C}^t_{j}\in\mc{C}^t|{\mc{C}_\text{seen}^{t-1}}_i\in{\mc{C}_\text{seen}^{t-1}})
    \\&=		
    \left\{\begin{aligned}
        &\frac{N_{ij}}{N_{i}},&\quad &\text{if CL-MLCL},\\
        &\frac{P({\mc{C}_\text{seen}^{t-1}}_i|\mc{C}^t_j)P(\mc{C}^t_j)}{P({\mc{C}_\text{seen}^{t-1}}_i)}=\frac{\mb{R}^t_{ij}N_j}{\sum_{\mb{x}}{\hat{z}}_{i}},&\quad &\text{if IL-MLCL},
      \end{aligned}
    \right.
  \end{aligned}
\end{equation}
where $N_{ij}$ is the accumulated number of examples with both class ${\mc{C}_\text{seen}^{t-1}}_i$ and $\mc{C}^t_{j}$, $N_{i}$ is the accumulated number of examples with class ${\mc{C}_\text{seen}^{t-1}}_i$.

{Finally, we construct an ACM using the soft and hard label statistics (IL-MLCL) or only the hard label statistics (CL-MLCL). Based on the established ACM, the GCN can capture the label dependencies across different tasks, improving the performance of continual multi-label recognition tasks.}

\begin{algorithm}[t]
  \caption{Training procedure of AGCN++.}
  \label{alg:main}
  \LinesNumbered
  \small\KwIn{$\mc{D}^t_\text{trn}$}
  \small
  \For{$t=1:T$}{
    \For{$(\mb{x},\mb{y}) \sim$ $\mc{D}^t_\text{trn}$}{
      \eIf{$t=1$}{
        Compute $\mb{A}^1$ with $\mb{y}$ using Eq.~\eqref{eq:hard-hard}\;
        $\mb{H}^1,~\mb{\hat{y}}_\text{cal} = \text{PLE}(\text{CNN}(x))$\;
        $\mb{\hat{y}}_\text{gph} = \text{GCN}(\mb{A}^1, \mb{H}^{1})$\;
        $\mb{\hat{y}} = \sigma\left( \mb{\hat{y}}_\text{cal} \oplus \mb{\hat{y}}_\text{gph}\right)$\;
        $\ell=\ell_\text{cls}(\mb{y},\mb{\hat{y}})$
      }{
        $\mb{\hat{z}}=\text{expert}(x)$\;  \tcp{\small get soft labels.}
        Compute $\mb{B}^t$ with $\mb{y}$ using Eq.~\eqref{eq:hard-hard};\\
        Compute $\mb{R}^t$ and $\mb{Q}^t$ using Eq.~\eqref{eq:soft-hard} and \eqref{eq:hard-soft}\;
        ${\mb{A}}^{t}=\begin{bmatrix} {\mb{A}}^{t-1} & \mb{R}^t \\ \mb{Q}^t & \mb{B}^t \end{bmatrix}$\;
        \tcp{\small compute ACM of task $t$.} 
        $\mb{H}^t,~\mb{\hat{y}}_\text{cal} = \text{PLE}(\text{CNN}(x))$\;
        $\mb{\hat{y}}_\text{gph} = \text{GCN}(\mb{A}^t, \mb{H}^{t})$\;
        \tcp{\small get new graph represnetation.}
        $\mb{y}^{\prime}_\text{gph}=\text{GCN}_\text{xpt}(\mb{A}^{t-1}, \mb{H}^{t-1})$\;
        \tcp{\small get target represnetation.} 
        $\mb{\hat{y}} = \sigma\left( \mb{\hat{y}}_\text{cal} \oplus \mb{\hat{y}}_\text{gph}\right)$\;

      \tcp{\small get class prediction.} 
        $\ell=\lambda_1\ell_\text{cls}(\mb{y},\mb{\hat{y}})+\lambda_2\ell_\text{dst}(\mb{\hat{z}},\mb{\hat{y}}_\text{old})$\\$~~~~~+\lambda_3\ell_\text{gph}(\mb{y}^{\prime}_\text{gph},\mb{\hat{y}}_\text{gph})$\;
        \tcp{\small compute the final loss.} 
      }
       Update $\text{AGCN++}$ model by minimizing $\ell$ \\
    }
    Update expert model \\
    \tcp{\small save parameters to the expert model.}
  }
\end{algorithm}

\subsection{Objective function}
\label{sec:loss}
As mentioned above, The class-incremental prediction scores $\mb{\hat{y}}$ for an image $\mb{x}$ can be calculated by Eq.~\eqref{eq:predict}.
The prediction $\mb{\hat{y}} = [\mb{\hat{y}}_\text{old}~\mb{\hat{y}}_\text{new}]\in\mbb{R}^{|\mc{C}_\text{seen}^{t}|}$, where $\mb{\hat{y}}_\text{old}\in\mbb{R}^{|\mc{C}_\text{seen}^{t-1}|}$ for old classes  and $\mb{\hat{y}}_\text{new}\in\mbb{R}^{|\mc{C}^t|}$ for new classes when $t>1$, $\mc{C}_\text{seen}^t=\mc{C}_\text{seen}^{t-1}\cup\mc{C}^t$. By binarizing the ground truth to hard labels $\mb{y} = [y_1, \cdots, y_{|\mc{C}|}], y_i\in\{0,1\}$, we train the current task using the Cross Entropy loss:
\begin{equation}\label{eq:current_loss}	
		\ell_\text{cls}(\mb{y},\mb{\hat{y}})=
		-\sum_{i=1}^{|\mc{C}|}\Big[y_i\log\left({{\hat{y}}}_{i}\right)+\left(1-{{y}}_{i}\right)\log\left(1-{{\hat{y}}}_{i}\right)\Big],
\end{equation}
where
$\mc{C}=\mc{C}_\text{seen}^t$ in CL-MLCL and $\mc{C}=\mc{C}^t$ in IL-MLCL.
However, like traditional SLCL, sequentially fine-tuning the model on the current task will lead to class-level forgetting of the old classes.
To mitigate the class-level catastrophic forgetting, based on the expert network, we construct the distillation loss as
\begin{equation}\label{eq:distillation}	\ell_\text{dst}({\mb{\hat{z}}},\mb{\hat{y}}_\text{old})=-\sum_{i=1}^{|\mc{C}_\text{seen}^{t-1}|}\left[{{\hat{z}_i}}\log\left({\hat{y}}_{i}\right)+\left(1-{{\hat{z}}}_{i}\right)\log\left(1-{\hat{y}}_{i}\right)\right],
\end{equation}
where ${\mb{\hat{z}}}$ is the soft labels used to represent the prediction on old classes.
The soft labels ${\mb{\hat{z}}}$ are used to be the target feature for the old class prediction $\mb{\hat{y}}_\text{old}$. 
Soft labels play two main roles in our paper: 1)  ${\mb{\hat{z}}}$ are used to be the target feature of old classes to mitigate the class-level forgetting in IL-MLCL and CL-MLCL scenarios; 2)  IL-MLCL has the past-missing partial label problem, soft labels are used as substitutes for old class labels to build label relationships across new and old classes.

To mitigate relationship-level forgetting across tasks, we constantly preserve the established relationships in the sequential tasks.
We compute the old graph representation to serve as a teacher to guide the training of the new GCN model.
The old graph representation in task $t$ is computed by $\mb{y}^{\prime}_\text{gph}=\text{GCN}_\text{xpt}(\mb{A}^{t-1}, \mb{H}^{t-1})$, $t>1$.
Then, we propose a relationship-preserving loss as a relationship constraint:
\begin{equation}\label{eq:graph_distillation}
		\ell_\text{gph}(\mb{y}^{\prime}_\text{gph},\mb{\hat{y}}_\text{gph})=\sum^{|\mc{C}_\text{seen}^{t-1}|}_{i=1} \left\Vert{\mb{y}^{\prime}_{\text{gph},i}-\mb{\hat{y}}_{\text{gph},i}}\right\Vert^2.
\end{equation}
By minimizing $\ell_\text{gph}$ with the partial constraint of old node embedding, the changes of $\text{GCN}$ parameters are limited.
Thus, the forgetting of the established label relationships is alleviated with the progress of MLCL image recognition. 

The final loss for the IL- and CL-MLCL model training is defined as
\begin{equation}\label{eq:final_loss}
	\ell=\lambda_1\ell_\text{cls}(\mb{y},\mb{\hat{y}})+\lambda_2\ell_\text{dst}(\mb{\hat{z}},\mb{\hat{y}}_\text{old})+ \lambda_3\ell_\text{gph}(\mb{y}^{\prime}_\text{gph},\mb{\hat{y}}_\text{gph}),
\end{equation}
where $\ell_\text{cls}$ is the classification loss, $\ell_\text{dst}$ is used to mitigate the class-level forgetting and $\ell_\text{gph}$ is used to reduce the  relationship-level forgetting. $\lambda_1$, $\lambda_2$ and $\lambda_3$ are the loss weights for $\ell_\text{cls}$, $\ell_\text{dst}$ and $\ell_\text{gph}$, respectively. 

{The AGCN++ algorithm for both IL-MLCL and CL-MLCL scenarios is presented in Algorithm~\ref{alg:main} to show the detailed training procedure.
Given the training dataset $\mc{D}_\text{trn}^t$:
(1) For the first task, the intra-task correlation matrix $\mb{A}^1$ is constructed by the statistics of hard labels $\mb{y}$, After the input $\mb{x}$ is fed to the CNN, the class-specific feature $\mb{\hat{y}}_\text{cal}$ and the graph nodes $\mb{H}^{1}$ is obtained by PLE.  Then the GCN encodes $\mb{A}^1$ and $\mb{H}^{1}$ to get graph representation $\mb{\hat{y}}_\text{gph}$. The prediction score $\mb{\hat{y}}$ is generated by $\mb{\hat{y}}_\text{cal}$  and $\mb{\hat{y}}_\text{gph}$ (Line 4-8).
(2) When $t>1$, the ACM $\mb{A}^t$ is augmented via soft labels $\mb{\hat{z}}$ and the Bayes' rule. Based on the $\mb{A}^t$, $\text{GCN}$ model can capture both intra- and inter-task label dependencies. Then, $\mb{\hat{z}}$  and $\mb{y}^{\prime}_\text{gph}$ as target features to build $\ell_\text{dst}$ and $\ell_\text{gph}$ respectively (Line 9-20).
(3) The AGCN++ and expert models are updated respectively (Line 21-23).}


\section{Experiments}

\begin{figure}
  \centering
  \includegraphics[width=\linewidth]{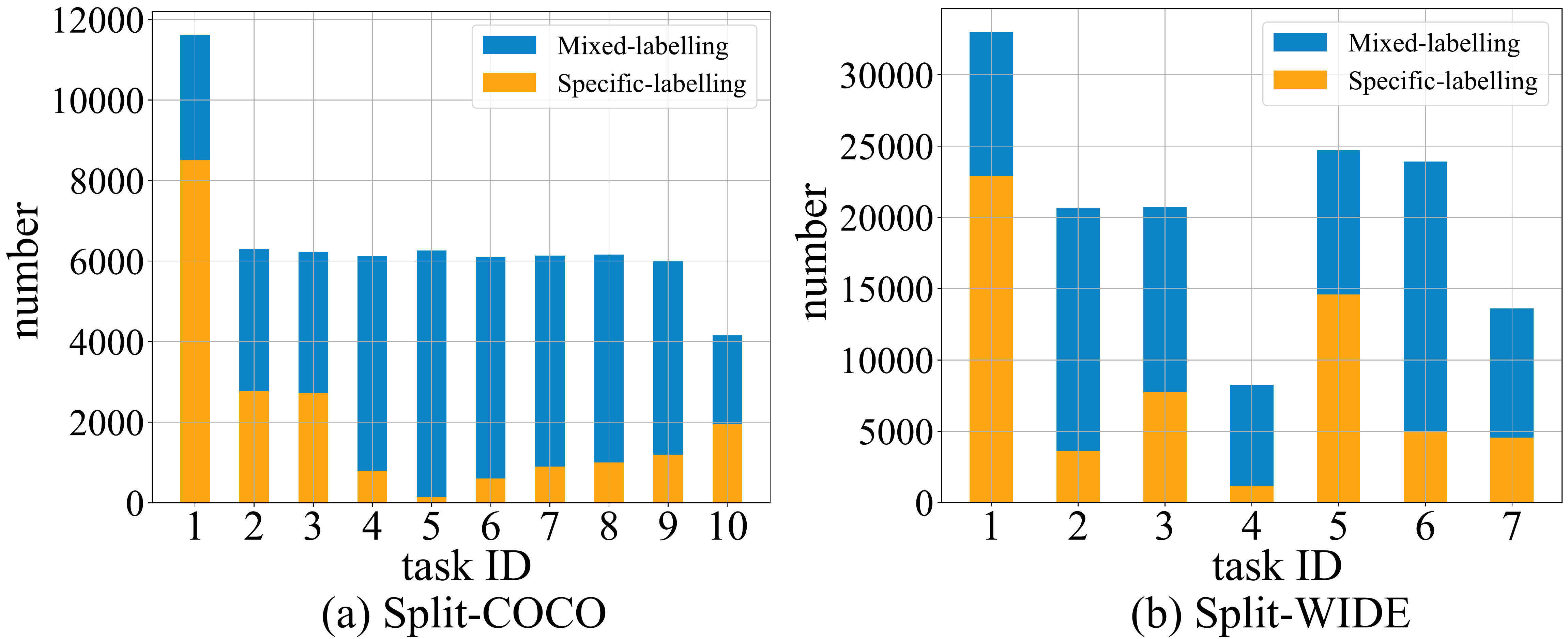}
  \caption{Dataset construction of Split-COCO and Split-WIDE.}
  \label{fig:dataset}
\end{figure}

\begin{table*}[t]
	\centering
	\caption{We report 7 metrics (\%) for multi-label classification after the whole data stream is seen once on \textbf{Split-WIDE} in both IL-MLCL and CL-MLCL scenarios. The Multi-Task is offline trained as the \textbf{upper bound}, and Fine-Tuning is the \textbf{lower bound}.} 
	\resizebox{.99\linewidth}{!}{
	\begin{tabular}{c|rrrrrrr|rrrrrrr}
		\hline
		\toprule
		\multirow{2}{*}{{Method}} 
		& \multicolumn{7}{c|}{\textbf{Split-WIDE IL-MLCL}}& \multicolumn{7}{c}{\textbf{Split-WIDE CL-MLCL}} \\
		\cline{2-15}
		&{mAP} $\uparrow$&{CP} $\uparrow$&{CR} $\uparrow$&{CF1} $\uparrow$&{OP} $\uparrow$&{OR} $\uparrow$&{OF1} $\uparrow$&{mAP} $\uparrow$&{CP} $\uparrow$&{CR} $\uparrow$&{CF1} $\uparrow$&{OP} $\uparrow$&{OR} $\uparrow$&{OF1} $\uparrow$ \\
		\hline
		\hline
		{Multi-Task}&66.17&69.15&55.30&61.45&77.74&66.30&71.57
		&69.19&60.60&43.96&50.33&78.45&59.39&66.67 \\
		\hline\hline
		{Fine-Tuning}&20.33&15.21&37.85&19.10&25.15&61.62&35.72
		&41.82&44.48&34.73&39.00&52.94&42.97&47.43 \\
		\rowcolor{mygray}
		Forgetting $\downarrow$&40.85&36.95&27.20&31.20&27.22&12.14&15.10
		&18.20&20.23&41.83&28.07&3.29&16.63&11.56 \\
		\hline
		{EWC}~\cite{kirkpatrick2017overcoming}&22.03&15.99&39.53&22.78&24.92&62.97&35.70
		&45.04&45.33&{37.13}&40.82&54.36&{53.81}&54.08 \\
				\rowcolor{mygray}
		Forgetting $\downarrow$&34.86&35.51&24.23&28.18&28.41&9.55&15.17
		&14.73&18.31&38.72&25.17&2.20&7.09&4.06 \\
		\hline
		{LwF}~\cite{li2017learning}&29.46&21.65&46.96&29.64&30.77&69.70&42.69
		&46.44&51.05&33.01&40.09&54.24&46.40&50.01 \\
				\rowcolor{mygray}
		Forgetting $\downarrow$&20.26&29.21&17.02&18.99&20.94&3.84&5.73
		&12.68&9.08&42.93&26.05&2.23&14.33&9.31 \\
		\hline
		{AGEM}~\cite{chaudhry2018efficient}&32.47&23.26&58.44&33.28&26.36&74.40&38.93
		&46.83&50.48&27.67&35.75&47.93&35.18&40.58 \\
				\rowcolor{mygray}

		Forgetting $\downarrow$&16.42&28.09&8.67&15.71&26.55&6.95&9.73
		 	&11.91&10.05&48.36&33.55&11.21&21.56&17.22\\
		\hline
		{ER}~\cite{rolnick2019experience}&34.03&24.64&60.02&34.94&26.62&75.57&39.37
		&48.08&54.33&31.16&39.61&53.40&38.84&44.98 \\
				\rowcolor{mygray}
		Forgetting $\downarrow$&15.15&26.18&7.14&11.80&26.45&6.25&8.61
		&9.24&7.13&45.32&27.58&2.96&19.28&14.53\\
		\hline
		{PRS}~\cite{kim2020imbalanced}&{39.70}&\textbf{52.77}&18.24&26.48&\textbf{60.81}&14.05&22.19
		&51.42&\textbf{58.26}&37.64&45.73&\textbf{55.66}&48.90&52.06 \\
				\rowcolor{mygray}
		Forgetting $\downarrow$&11.24&4.08&43.22&34.48&2.34&55.73&43.76
		&7.86&2.21&37.12&16.68&1.90&11.36&7.13 \\
		\hline
		
				{SCR}~\cite{mai2021supervised}&35.34&28.33&54.34&35.47&32.21&70.28&41.92
		&49.23&53.87&36.86&43.77&50.16&47.58&48.84  \\
				\rowcolor{mygray}
		Forgetting $\downarrow$&14.26&21.29&9.56&10.17&23.09&7.26&8.04
		&8.34&7.89&39.22&20.56&6.62&13.56&10.78 \\
		\hline
		{AGCN}&{42.15}&26.04&\textbf{70.21}&{37.99}&29.53&\textbf{84.02}&{43.70}
		&{54.20}&56.24&{39.10}&{46.13}&53.94&{56.84}&{55.35} \\
				\rowcolor{mygray}
		Forgetting $\downarrow$&10.34&25.44&1.35&5.82&25.23&1.12&4.12
		&5.27&4.56&34.54&14.34&2.49&5.40&3.26 \\
		\hline

		\textbf{AGCN++}&\textbf{45.73}&33.08&{61.57}&\textbf{43.04}&32.29&{75.62}&\textbf{45.26}
		&\textbf{57.07}&55.07&\textbf{54.26}&\textbf{54.66}&49.03&\textbf{74.96}&\textbf{59.29} \\
				\rowcolor{mygray}
		Forgetting $\downarrow$&8.32&17.28&7.44&2.13&22.11&5.52&3.98
		&4.45&3.56&19.48&10.64&6.13&0.18&1.02 \\
		\bottomrule

	\end{tabular}}
\label{tab:results_WIDE}
\end{table*}

\begin{table*}[t]
	\centering
	\caption{We report seven metrics (\%) for multi-label classification after the whole data stream is seen once on \textbf{Split-COCO} in both IL-MLCL and CL-MLCL scenarios. The Multi-Task is offline trained as the \textbf{upper bound}, and Fine-Tuning is the \textbf{lower bound}.} 

	\resizebox{.99\linewidth}{!}{
	\begin{tabular}{c|rrrrrrr|rrrrrrr}
		\hline
		\toprule
		\multirow{2}{*}{{Method}} 
		& \multicolumn{7}{c|}{\textbf{Split-COCO IL-MLCL}}& \multicolumn{7}{c}{\textbf{Split-COCO CL-MLCL}} \\
		\cline{2-15}
		&{mAP} $\uparrow$&{CP} $\uparrow$&{CR} $\uparrow$&{CF1} $\uparrow$&{OP} $\uparrow$&{OR} $\uparrow$&{OF1} $\uparrow$&{mAP} $\uparrow$&{CP} $\uparrow$&{CR} $\uparrow$&{CF1} $\uparrow$&{OP} $\uparrow$&{OR} $\uparrow$&{OF1} $\uparrow$ \\
		\hline
		\hline
		{Multi-Task}&65.85&71.64&54.31&61.79&77.24&58.03&66.27
		&68.33&73.06&49.82&61.49&89.31&64.99&74.98\\
		\hline\hline
		{Fine-Tuning}&9.83&6.90&18.52&10.54&21.63&41.25&28.83
		&32.35&	33.34&29.10&31.01&57.03&45.56&50.56
		 \\
		 \rowcolor{mygray}
		Forgetting $\downarrow$&58.04&48.96&64.30&63.54&18.24&38.76&20.60
		&31.78&33.32&32.14&33.90&16.61&9.22&12.24 \\
		\hline
		{EWC}~\cite{kirkpatrick2017overcoming}&12.20&9.70&17.54&12.50&23.63&39.84&29.67
		&35.83&	31.88&33.05&32.18&57.62&46.98&51.60 \\
				\rowcolor{mygray}
		Forgetting $\downarrow$&45.61&42.68&60.50&55.44&15.34&40.59&19.85
		&27.66&37.82&26.29&30.29&16.24&7.18&10.16 \\
		\hline
		{LwF}~\cite{li2017learning}&19.95&18.02&28.44&21.69&33.14&57.83&40.68
		&40.87&	44.36&{35.07}&39.07&61.72&48.10&53.95 \\
				\rowcolor{mygray}
		Forgetting $\downarrow$&41.16&29.73&44.01&39.85&8.70&19.38&11.43
		&21.15&22.70&25.90&23.64&12.29&4.99&7.67 \\
		\hline
		{AGEM}~\cite{chaudhry2018efficient}&23.31&22.34&42.10&27.25&29.95&62.32&37.94
		&42.25&	64.40&19.28&29.08&57.64&12.62&18.59 \\
				\rowcolor{mygray}
		Forgetting $\downarrow$&34.52&17.12&20.36&18.92&13.02&11.35&12.94
		&19.75&9.11&45.66&35.37&15.94&34.80&39.92 \\
		\hline
		{ER}~\cite{rolnick2019experience}&25.03&26.45&41.14&30.54&30.32&61.84&38.38
		&43.54&\textbf{71.15}&16.65&26.60&62.89&17.72&26.44 \\
				\rowcolor{mygray}
		Forgetting $\downarrow$&33.46&14.96&22.28&17.28&11.80&12.49&12.34
		&17.13&1.14&47.73&38.34&11.79&30.32&32.66 \\
		\hline
		
		{PRS}~\cite{kim2020imbalanced}&{31.08}&\textbf{56.07}&22.74&32.27&\textbf{57.87}&14.24&22.25
		&46.39&58.56&29.41&38.20&58.84&{50.54}&54.25 \\
				\rowcolor{mygray}
		Forgetting $\downarrow$&28.82&1.32&50.59&16.21&0.34&58.37&30.43
		&13.07&13.52&31.23&24.56&14.21&6.09&6.36 \\
		\hline

		{SCR}~\cite{mai2021supervised}&25.75&25.22&{49.35}&30.63&29.40&{69.91}&39.10
		&44.96&54.00&29.25&37.82&41.47&40.40&40.46 \\
				\rowcolor{mygray}
		Forgetting $\downarrow$&32.02&15.27&16.02&15.98&13.58&6.52&11.96
		&15.33&19.88&31.89&25.12&30.04&13.24&19.26 \\
		\hline
		{AGCN}&{34.11}&31.80&{47.73}&{35.49}&34.38&{67.72}&{42.37}   
		&{48.82}&55.73&30.83&{39.18}&\textbf{74.27}&47.06&{56.76} \\
				\rowcolor{mygray}
		Forgetting $\downarrow$&23.71&12.21&17.81&14.79&8.03&9.86&8.16
		&10.40&18.72&30.39&22.38&1.63&6.71&3.97 \\
		\hline

				\textbf{AGCN++}&\textbf{38.23}&32.51&\textbf{60.47}&\textbf{41.38}&34.32&\textbf{75.34}&\textbf{45.26}
		&\textbf{53.49}&46.66&\textbf{52.96}&\textbf{49.55}&55.14&\textbf{64.74}&\textbf{59.32} \\
				\rowcolor{mygray}
		Forgetting $\downarrow$&20.12&11.13&9.08&11.34&8.82&3.68&6.78
		&7.24&24.88&6.34&14.52&23.67&1.34&1.24 \\
		\bottomrule
		
	\end{tabular}}
\label{tab:results_COCO}
\end{table*}

\begin{figure}
	\centering
	\includegraphics[width=\linewidth]{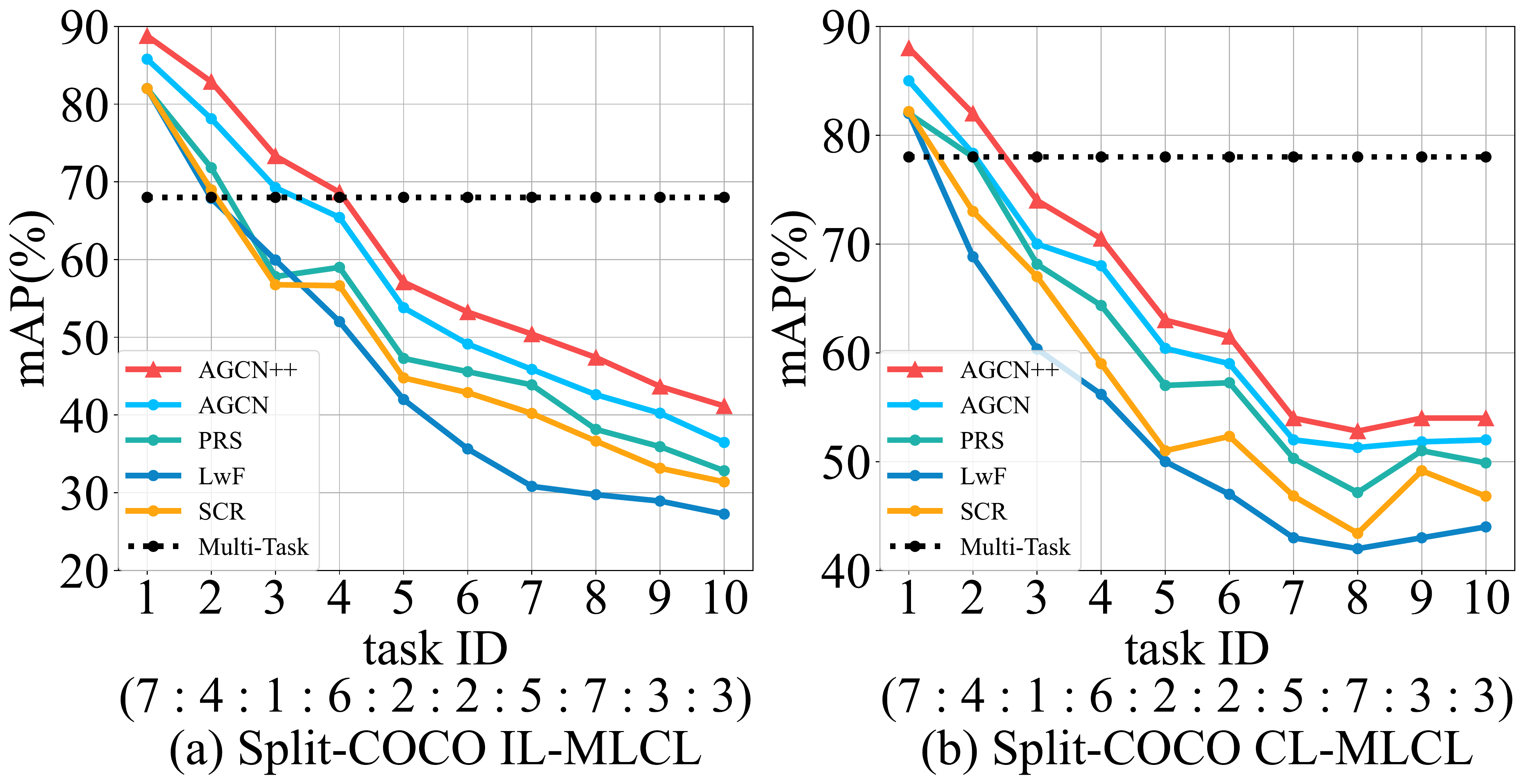}
	\caption{mAP (\%) of different classes setting in both IL- and CL-MLCL.}
	\label{fig:different_class}
  \end{figure}

\begin{figure*}
  \centering
  \includegraphics[width=\linewidth]{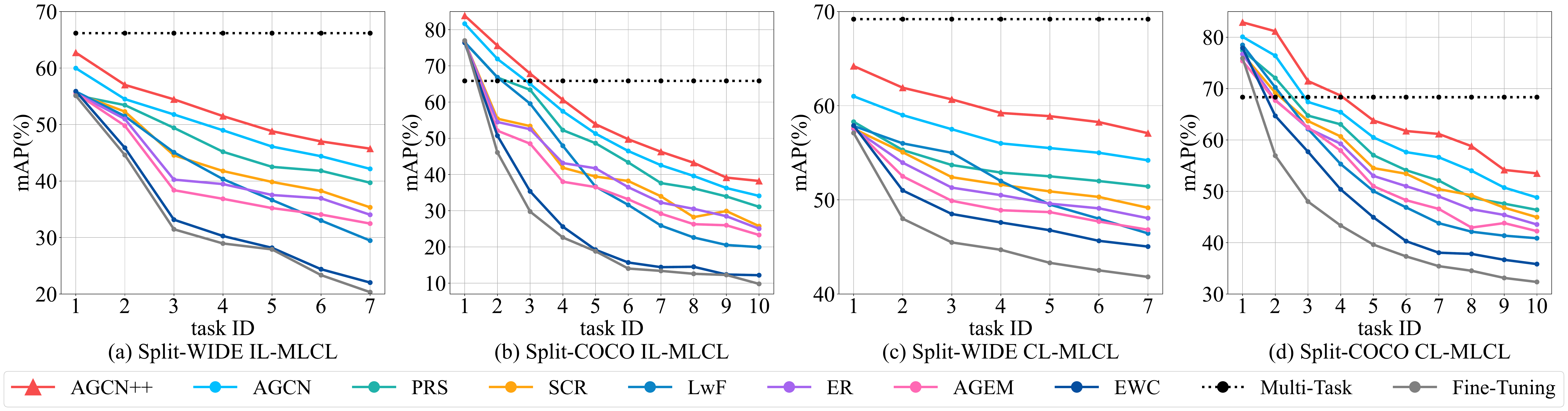}
  \caption{{mAP (\%) on two benchmarks in both IL- and CL-MLCL.}}
  \label{fig:learning_curves}
\end{figure*}

\subsection{Datasets}

\subsubsection{Dataset description}

We use two datasets, Split-COCO and Split-WIDE, to evaluate the effectiveness of the proposed method.

\textbf{Split-COCO}.
We choose the 40 most frequent concepts from 80 classes of MS-COCO~\cite{lin2014microsoft} to construct Split-COCO, which has 65082 examples for training and 27,173 examples for validation.
The 40 classes are split into ten different and non-overlapping tasks, each containing four classes.

\textbf{Split-WIDE}. 
NUS-WIDE~\cite{chua2009nus} is a raw web-crawled multi-label image dataset. We further curate a sequential class-incremental dataset from NUS-WIDE.  Following~\cite{jiang2017deep}, we choose the 21 most frequent concepts from 81 classes of NUS-WIDE to construct the Split-WIDE, which has 144,858 examples for training and 41,146 examples for validation. Split-WIDE has a larger scale than Split-COCO.
We split the Split-WIDE into 7 tasks, where each task contains 3 classes.

\subsubsection{Dataset collection}

{We enlist the curation details of Split-COCO and Split-WIDE.
In the previous continual learning methods, Shmelkov \etal\cite{shmelkov2017incremental} selects 20 out of 80 classes to create 2 tasks for SLCL, each with 10 classes.
Nguyen \etal\cite{nguyen2019contcap} tailor MSCOCO for continual learning of captioning. They select 24 out of 80 classes to create two tasks.
PRS~\cite{kim2020imbalanced} needs more low-frequency classes to study the imbalanced problem. They curate four tasks with 70 classes using MSCOCO.
Compared to these previous splitting, on the one hand, we set more tasks to test the robustness of the algorithm over more tasks to create a continual setting.
On the other hand, we selected more frequent concepts from the original dataset to reduce the long-tail effect of the original data.
Multi-label datasets inherently have intersecting concepts among the data points. Hence, a naive splitting strategy may lead to a dangerous amount of data loss. This motivates us to minimize data loss during the split. 
Moreover, to test diverse research environments, the second objective is to keep the size of the splits balanced optionally.  
To split the well-known MS-COCO and NUS-WIDE into several different tasks fairly and uniformly, we introduce two kinds of labelling in the datasets. 
\textit{1) Specific-labelling}: If an image only has the labels that belong to the task-special class set $\mc{C}^t$ of task $t$, we regard it as a specific-labelling image for task $t$;
\textit{2) Mixed-labelling}: If an image not only has the task-specific labels but also has the old labels belonging to the class set $\mc{C}^{t-1}_{\text{seen}}$, we regard it as a mixed-labelling image.}

In IL-MLCL, because the model learns from the task-specific labels $\mc{C}^t$, the training data is labelled without old labels, so IL-MLCL will suffer from the partial label problem, which mainly appears in the mixed-labelling image. The IL-MLCL and CL-MLCL share the same training images. 
A randomly data-splitting approach may lead to the imbalance of specific-labelling and mixed-labelling images for each task.
We split two datasets into sequential tasks with the following strategies to ensure a proper proportion.
We first count the number of labels for each image.
Then, we give priority to leaving specific-label images for each task. The mixed-labelling images are then allocated to other tasks. The dataset construction is presented in Fig~\ref{fig:dataset}. 

\subsection{Evaluation metrics}

\noindent
\textbf{Multi-label evaluation}. Following these multi-label learning methods~\cite{chen2019multi,chen2020knowledge,chen2021learning}, 7 metrics are leveraged in MLCL.
(1) the average precision (AP) on each label and the mean average precision (\textbf{mAP}) over all labels;
(2) the per-class F1-measure (\textbf{CF1});
(3) the overall F1-measure (\textbf{OF1}). The mAP, CF1 and OF1 are relatively more important for multi-label performance evaluation.
Moreover, we adopt 4 other metrics: per-class precision (CP), per-class recall (CR), overall precision (OP) and overall recall (CR).
\begin{equation*}
	\label{eq:metrics}
	\begin{aligned}
		& \text{OP}=\frac{\sum_iN_i^c}{\sum_iN_i^p},~~~~~~~ 
		~~~~~~~~\text{CP}=\frac{1}{C}\sum_i\frac{N_i^c}{N_i^p},\\
		& \text{OR}=\frac{\sum_iN_i^c}{\sum_iN_i^g},~~~~~~~ 
		~~~~~~~~\text{CR}=\frac{1}{C}\sum_i\frac{N_i^c}{N_i^g},\\
		& \text{OF1}=\frac{2 \times \text{OP} \times \text{OR}}{\text{OP}+\text{OR}},~ 
		\text{CF1}=\frac{2 \times \text{CP} \times \text{CR}}{\text{CP}+\text{CR}},	
	\end{aligned}
\end{equation*}
where $i$ is the class label and $C$ is the number of labels. $N_i^c$ is the number of correctly predicted images for class $i$, $N^p_i$ is the number of predicted images for class $i$ and $N_i^g$ is the number of ground-truth for class $i$. 


\noindent
\textbf{Forgetting measure}~\cite{chaudhry2018riemannian}. 
This metric denotes the above multi-label metric value difference for each task between testing when it was first trained, and the last task was trained.
For example, the forgetting measure of mAP for a task $t$ can be computed by its performance difference between task $T$ and $t$ was trained. $F_{t}$, average forgetting after the model has been trained continually  up till task $t\in\{1,\cdots,T\}$ is defined as:
\begin{equation} \label{eq:forget}
	F_{t}=\frac{1}{t-1} \sum_{j=1}^{t-1} f_{j}^{t},
\end{equation}
where $f_{j}^{t}$ is the forgetting on task $j$ after the model is trained up till task $t$ and computed as
\begin{equation}
	f_{j}^{t}=\max _{l \in\{1, \cdots, k-1\}} a_{l, j}-a_{t, j},
\end{equation}
where $a$ denotes every metric in MLCL like mAP, CF1 and OF1. We evaluate the final forgetting ($F_{T}$) after training the final task. 

\begin{table*}[t]
	\centering
	\caption{We report 3 more important metrics (\%) for multi-label classification after the whole data stream is seen once on 8-way Split-COCO and 7-way Split-WIDE in both IL-MLCL and CL-MLCL scenarios. }
	\resizebox{.85\linewidth}{!}{
	\begin{tabular}{c|rrr|rrr|rrr|rrr}
		\toprule
		
		\multirow{2}{*}{{Method}} 
        & \multicolumn{3}{c|}{{Split-COCO (IL 8-way)}} & \multicolumn{3}{c|}{{Split-COCO (CL 8-way)}} & \multicolumn{3}{c|}{{Split-WIDE (IL 7-way)}} & \multicolumn{3}{c}{{Split-WIDE (CL 7-way)}} \\ \cline{2-13} 
		&{mAP}&{CF1}&{OF1}&{mAP}&{CF1}&{OF1}&{mAP}&{CF1}&{OF1}&{mAP}&{CF1}&{OF1} \\
		\hline\hline
		{Multi-Task}&72.24&64.34&72.35
		&74.63&68.87&79.74&64.71&58.86&53.82
		&68.84&58.64&60.23 \\
		\hline\hline
		{Fine-Tuning}&16.89&8.82&46.63
		&62.69&56.87&68.45&39.26&34.51&50.10
		&55.46&41.76&57.01 \\

		 \hline

		{EWC}~\cite{kirkpatrick2017overcoming}&23.47&11.03&47.82
		&63.10&56.91&68.60&39.56&33.20&47.69
		&56.73&{51.46}&66.07 \\

		 \hline
		{LwF}~\cite{li2017learning}&{36.67}&{37.38}&{46.98}
		&63.66&58.39&69.74&{40.14}&{43.64}&{56.22}
		&57.48&{54.90}&68.24 \\

		 \hline
		{AGEM}~\cite{chaudhry2018efficient}&40.43&33.50&45.75
		&63.92&55.30&69.34&42.33&40.66&52.94
		&57.52&45.36&58.25 \\

		 \hline
		{ER}~\cite{rolnick2019experience}&42.23&39.89&46.24
		&{64.22}&{58.80}&{70.29} &46.39&47.63&58.71
		&58.12&47.11&55.30\\

		 \hline
		 {PRS}~\cite{kim2020imbalanced}&{61.56}&47.81&33.09
		&{66.61}&{60.13}&{70.89} 
		&{47.45}&44.71&60.55
		&{58.15}&54.39&{68.92}\\

		 \hline
		{AGCN}&{{62.60}}&{{57.85}}&{{59.29}}
		&{{70.24}}&{{63.12}}&{{74.21}}  &{{54.11}}&{{53.71}}&{{67.04}}   
		&{{58.98}}&{{55.32}}&{{69.28}}\\

		\hline
		
		\textbf{AGCN++}&\textbf{{65.92}}&\textbf{{60.02}}&\textbf{{68.82}}
		&\textbf{{73.41}}&\textbf{{68.55}}&\textbf{{74.54}}  &\textbf{{56.63}}&\textbf{{57.51}}&\textbf{{73.58}}   
		&\textbf{{61.04}}&\textbf{{60.22}}&\textbf{{72.40}}\\

		\bottomrule
	\end{tabular}}
\label{tab:more_class}
\end{table*}

\begin{table*}
    \centering
    \caption{Ablation studies (\%) for  ACM $\mb{A}^t$ and $\textbf{PLE}$ on Split-WIDE and Split-COCO.
    }
    \resizebox{\linewidth}{!}{
        \begin{tabular}{c|cc|cccccc|cccccc}
        \toprule
          
           		\multicolumn{3}{c}{}&\multicolumn{6}{c}{Split-WIDE}&\multicolumn{6}{c}{Split-COCO}\\
		\multicolumn{3}{c}{}&\multicolumn{3}{c}{AGCN++ (w/ PLE)}&\multicolumn{3}{c}{AGCN (w/o PLE)}&\multicolumn{3}{c}{AGCN++ (w/ PLE)}&\multicolumn{3}{c}{AGCN (w/o PLE)}\\
           &$\mb{A}^{t-1}$ \& $\mb{B}^t$  & $\mb{R}^t$ \& $\mb{Q}^t$  
           &{mAP} $\uparrow$&{CF1} $\uparrow$&{OF1} $\uparrow$
           &{mAP} $\uparrow$&{CF1} $\uparrow$&{OF1} $\uparrow$
           &{mAP} $\uparrow$&{CF1} $\uparrow$&{OF1} $\uparrow$  
  &{mAP} $\uparrow$&{CF1} $\uparrow$&\textbf{OF1} $\uparrow$\\
           
            \hline
            \multirow{2}{*}{IL-MLCL}
           & $\surd$ & $\times$ 
           &  42.25 & 40.47 & 42.98
           &  38.05 & 34.03 & 42.71   
           &  35.19 & 39.98 & 37.62 
           &  31.52 & 30.37 & 34.87  \\
           & $\surd$ & $\surd$ 
           & \textbf{45.73} & \textbf{43.04} & \textbf{45.26}
           & \textbf{42.15} & \textbf{37.99} & \textbf{43.70}    
           & \textbf{38.23} & \textbf{41.38} & \textbf{45.26}
           & \textbf{34.11} & \textbf{35.49} & \textbf{42.37}\\
           	\hline
            \multirow{2}{*}{CL-MLCL}
           & $\surd$ & $\times$ 
           &  54.72 & 50.41 & 48.84
           &  49.47 & 44.73 & 52.13   
           &  51.51 & 47.13 & 56.97
           &  44.53 & 35.55 & 53.57\\
       
           & $\surd$ & $\surd$ 
           & \textbf{57.07} & \textbf{54.66} & \textbf{59.29}
           & \textbf{54.20} & \textbf{46.13} & \textbf{55.35}  
           & \textbf{53.49} & \textbf{49.55} & \textbf{59.32}
            & \textbf{48.82} & \textbf{39.18} & \textbf{56.76}\\
            \bottomrule
    \end{tabular}}
    \label{tab:ACM_ab}
\end{table*}

\subsection{Implementation details} 
Following existing multi-label image classification methods ~\cite{chen2019multi,chen2020knowledge,chen2021learning}, we employ ResNet101~\cite{he2016deep} as the image feature extractor pre-trained on ImageNet~\cite{deng2009imagenet}. We adopt Adam~\cite{kingma2014adam} as the optimizer of network with $\beta_1=0.9$, $\beta_2=0.999$, and $\epsilon=10^{-4}$. Following~\cite{chen2019multi,chen2020knowledge}, our AGCN++ consists of two GCN layers with output dimensionality of 1024 and 2048, respectively. The input images are randomly cropped and resized to $448\times448$ with random horizontal flips for data augmentation. The network is trained for a single epoch like most continual learning methods done~\cite{de2021continual,bang2021rainbow,lyu2020multi,rebuffi2017icarl}. 

\subsection{Baseline methods}
MLCL is a new paradigm of continual learning. We compare our method with several essential and state-of-art continual learning methods, including
(1) \textit{EWC}~\cite{kirkpatrick2017overcoming}, which regularizes the training loss to avoid catastrophic forgetting;
(2) \textit{LwF}~\cite{li2017learning}, which uses the distillation loss by saving task-specific parameters;
(3) \textit{ER}~\cite{rolnick2019experience}, which saves a few training data from the old tasks and retrains them in the current training;
(4) \textit{AGEM}~\cite{chaudhry2018efficient} resets the training gradient by combining the gradient on the memory and training data; 
(5) \textit{PRS}~\cite{kim2020imbalanced}, which uses an improved reservoir sampling strategy to study the imbalanced problem.
PRS studies similar problems with us. Still, they focus more on the imbalanced problem but ignore the label relationships and the problem of partial labels for MLCL image recognition.
(6) \textit{SCR}~\cite{mai2021supervised}, which proposes the NCM classifier to improve SLCL performance. SCR is an algorithm designed to improve the top-1 accuracy of single-label recognition.
Similar to~\cite{kim2020imbalanced,mai2021supervised,zhou2022few}, we use a \textit{Multi-Task} baseline, which is trained on a single pass over shuffled data from all tasks. It can be seen as the performance upper bound. We also compare with the \textit{Fine-Tuning}, which performs training without any continual learning technique. Thus, it can be regarded as the performance lower bound.
Note that, to extend some SLCL methods to MLCL, we turn the final Softmax layer in each of these methods into a Sigmoid. Other details follow their original settings.

\subsection{Main results}

\subsubsection{Split-WIDE results}
In Table~\ref{tab:results_WIDE}, with the establishment of relationships and inhibition of class-level and relationship-level forgetting using distillation and relationship-preserving loss, our method shows better performance than the other state-of-art performances in both IL-MLCL and CL-MLCL scenarios.
In particular, AGCN and AGCN++ perform better than other comparison methods on three more essential evaluation metrics, including mAP, CF1 and OF1, which means the effectiveness in multi-label classification.
Also, in the forgetting value evaluated after task $T$, we achieve a better forgetting measure, which means the stability of the proposed method in MLCL. 
In the IL-MLCL scenario, because we use soft labels to replace hard labels in the old task label space and establish and remember the label relationships, the AGCN++ outperforms the most state-of-art performances by a large margin: 45.73\% vs. 42.15\% (+ 3.58\%) on mAP, 43.04\% vs. 37.99\% (+ 5.05\%) on CF1 and 45.26\% vs. 43.70\%   (+ 1.56\%) on OF1, as shown in Table~\ref{tab:results_WIDE}. 
Like IL-MLCL, CL-MLCL still needs to model complete label dependencies between label relationships and reduce forgetting. 
The AGCN++ shows better performance than the others in CL-MLCL: 57.07\% vs. 54.20\% (+ 2.87\%) on mAP, 54.66\% vs. 46.13\% (+ 8.53\%) on CF1 and 59.29\% vs. 55.35\% (+ 3.94\%) on OF1, as demonstrated in Table~\ref{tab:results_WIDE}, which suggests that AGCN++ is effective in a large-scale multi-label dataset.


\subsubsection{Split-COCO results}
Split-COCO is split into ten tasks, as mentioned in~\cite{delange2021continual}, compared with methods PRS~\cite{kim2020imbalanced} and ER~\cite{rolnick2019experience}, our approach can protect privacy better because AGCN++ does not collect data from the original dataset. As shown in Table~\ref{tab:results_COCO}, in IL-MLCL and CL-MLCL, AGCN++ achieves better performance than the others in most metrics. 
AGCN++ also has a low rate of forgetting old knowledge.
With the AGCN++ combining  intra- and inter-task label relationships, the proposed AGCN++ outperforms the most state-of-art performances in IL-MLCL: 38.23\% vs. 34.11\% (+ 4.12\%) on mAP, 41.38\% vs. 35.49\% (+ 5.89\%) on CF1 and 45.26\% vs. 42.37\ (+ 2.89\%) on OF1. This means soft labels can effectively replace hard labels in the old task label space to alleviate the partial label problem.
AGCN++ is also better in CL-MLCL: 53.49\% vs. 48.82\% (+ 4.67\%) on mAP, 49.55\% vs. 39.18\% (+ 10.37\%) on CF1 and 59.32\% vs. 56.76\%  (+ 2.56\%) on OF1.
This means ACM is effective for both IL-MLCL and CL-MLCL scenarios in Split-COCO. As illustrated above, AGCN++ can be a uniform MLCL method for IL-MLCL and CL-MLCL.
\subsection{More MLCL settings}
In order to prove the robustness of the method, we verify the effectiveness of AGCN and AGCN++ under other MLCL settings. 
{First, as shown in Table~\ref{tab:more_class}, we increase the number of classes of each task to verify that the proposed PLE and ACM can effectively handle more label relationships in a task. Specifically, 8-way for Split-COCO and 7-way for Split-WIDE. As shown in Table~\ref{tab:more_class}, our AGCN and AGCN++ can still achieve better results in three more important metrics, mAP, CF1 and OF1. Take the mAP, for example. For Split-COCO, 65.92\% vs. 62.60\% (+ 3.32\%) in IL 8-way and 73.41\% vs. 70.24\% (+ 3.17\%) in CL 8-way. For Split-WIDE, 56.63\% vs. 54.11\% (+ 2.52\%) in IL 7-way and 61.04\% vs. 58.98\% (+ 2.06\%) in CL 7-way.}   
Second, considering in the real world, different tasks often have different numbers of classes. So we provide a different number of classes for each task in a random manner. Specifically, the task setting is "7 : 4 : 1 : 6: 2 : 2 : 5 : 7 : 3 : 3". As shown in Fig~\ref{fig:different_class}, our method performs better than other comparison methods in every task.
These experiments can prove the effectiveness of AGCN++ from more angles.

\subsection{mAP curves}
Similar to~\cite{mai2021supervised,bang2021rainbow,zhou2022few}, we show the mAP trends of different methods in Fig.~\ref{fig:learning_curves} for sequential learning. 
These curves indicate the performance along the MLCL progress.
In two MLCL scenarios, Fig.~\ref{fig:learning_curves} illustrates the mAP changes as tasks are being learned on two benchmarks.
The mAP curves show that AGCN and AGCN++ can perform better through the MLCL process.
In addition, their algorithm is applied after the first task for most continual learning methods. Our AGCN  and AGCN++ has modelled the label dependencies from the first task. As distillation loss and relationship-preserving loss are applied to subsequent tasks, the algorithm's performance exceeds other methods in each task.


\begin{table*}[t] 
    \centering
    \caption{Ablation studies (\%) for loss weights and relationship-preserving loss on Split-WIDE and Split-COCO for IL- and CL-MLCL.}
    \resizebox{0.86\linewidth}{!}{
        \begin{tabular}{c|ccc|ccc|ccc|ccc}
            \toprule
           &\multicolumn{6}{c}{Split-WIDE}&\multicolumn{6}{c}{Split-COCO}\\
         & $\lambda_1$ & $\lambda_2$   &$\lambda_3$   &{mAP} $\uparrow$&{CF1} $\uparrow$&{OF1} $\uparrow$ 
         & $\lambda_1$ & $\lambda_2$   &$\lambda_3$   &{mAP} $\uparrow$&{CF1} $\uparrow$&{OF1} $\uparrow$\\
            \hline

             \multirow{4}{*}{IL-MLCL}
             &$0.10$ & $0.90$   & $0$    & 42.04 & 38.76 & 42.12
             &$0.15$ & $0.85$   & $0$    & 36.77 & 39.95 & 39.10 
               \\
               &  \multicolumn{3}{c|}{\cellcolor{mygray}Forgetting $\downarrow$}    & \cellcolor{mygray}10.48 & \cellcolor{mygray}5.24 & \cellcolor{mygray}6.42
             &  \multicolumn{3}{c|}{\cellcolor{mygray}Forgetting $\downarrow$}    & \cellcolor{mygray}22.34 & \cellcolor{mygray}12.73 & \cellcolor{mygray}13.32  
               \\

              \cline{2-13}
              &$0.10$ & $0.90$   & $ 10^4 $    & \textbf{45.73} & \textbf{43.04} & \textbf{45.26}
           &$0.15$ & $0.85$   & $ 10^4 $    & \textbf{38.23} & \textbf{41.38} & \textbf{45.26}   
            \\
             &\multicolumn{3}{c|}{\cellcolor{mygray}Forgetting $\downarrow$}    & \cellcolor{mygray}8.32 & \cellcolor{mygray}2.13 & \cellcolor{mygray}3.98
             &\multicolumn{3}{c|}{\cellcolor{mygray}Forgetting $\downarrow$}    & \cellcolor{mygray}20.12 & \cellcolor{mygray}11.34 & \cellcolor{mygray}6.78 
             \\
			
            \bottomrule
            
            \multirow{4}{*}{CL-MLCL}

              &$0.70$ & $0.30$   & $0$    & 55.68 & 51.58 & 49.24
             &$0.40$ & $0.60$   & $0$    & {50.98} & 46.82 & 54.70
                \\
                &  \multicolumn{3}{c|}{\cellcolor{mygray}Forgetting $\downarrow$}    &\cellcolor{mygray}5.04  &\cellcolor{mygray}12.56  &\cellcolor{mygray}10.24
              &  \multicolumn{3}{c|}{\cellcolor{mygray}Forgetting $\downarrow$}    & \cellcolor{mygray}12.87 & \cellcolor{mygray}22.80 & \cellcolor{mygray}6.62  
                  \\

              \cline{2-13}

             &$0.70$ & $0.30$   & $ 10^3 $    & \textbf{57.07} & \textbf{54.66} & \textbf{59.29}
             &$0.40$ & $0.60$   & $ 10^4 $    & \textbf{53.49} & \textbf{49.55} & \textbf{59.32}   
             \\
             &\multicolumn{3}{c|}{\cellcolor{mygray}Forgetting $\downarrow$}    & \cellcolor{mygray}4.45 &\cellcolor{mygray} 10.64 & \cellcolor{mygray}1.02
             &\multicolumn{3}{c|}{\cellcolor{mygray}Forgetting $\downarrow$}    & \cellcolor{mygray}7.24 & \cellcolor{mygray}14.52 & \cellcolor{mygray}1.24 
              \\

             \bottomrule
    \end{tabular}}
    \label{tab:weight}
\end{table*}

\begin{figure*}[t]
	\centering
	\includegraphics[width=0.9\linewidth]{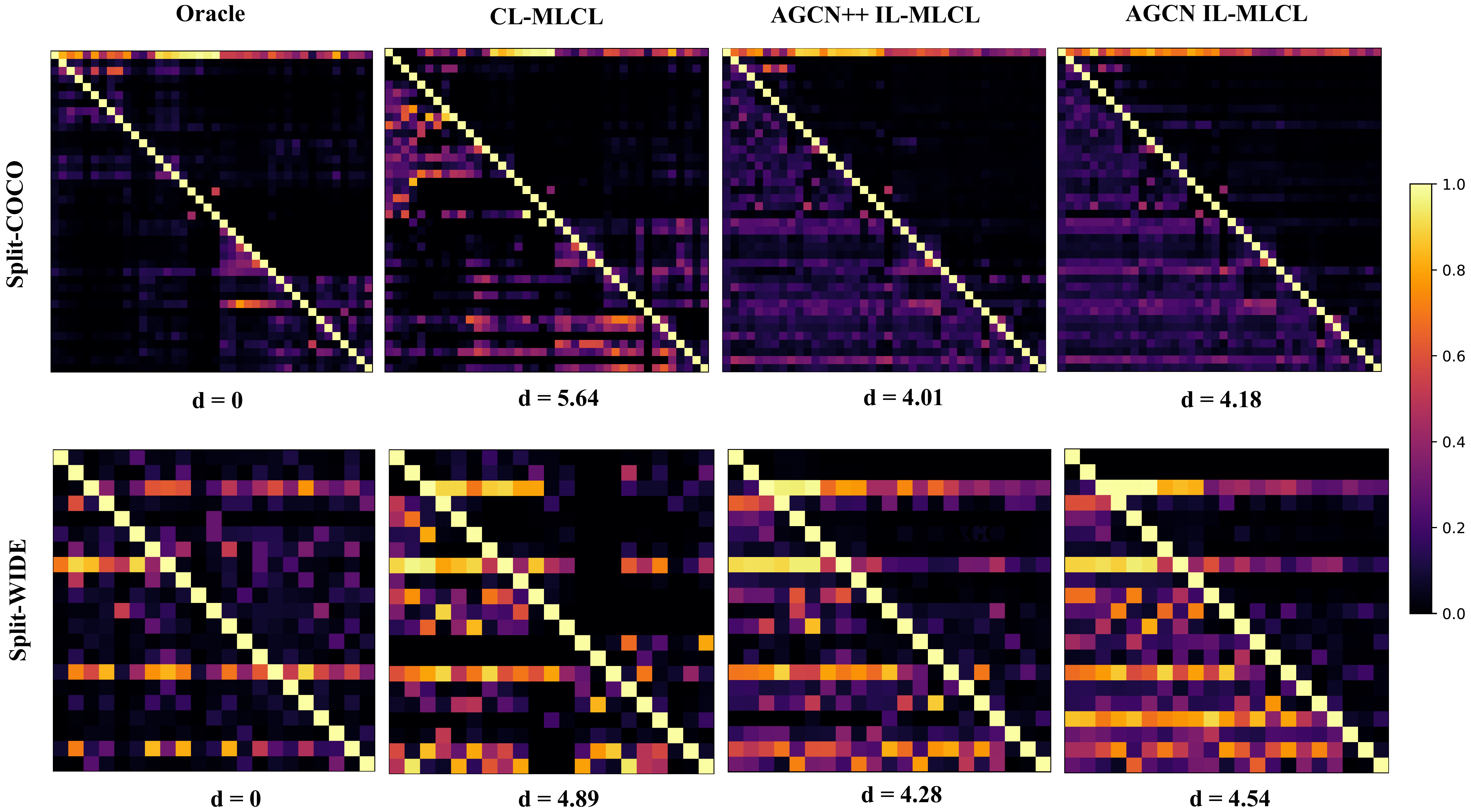}
	\caption{ACM visualization. It shows the intra- and inter-task label relationships are constructed well.}
	\label{fig:ACM_visual}
\end{figure*}

\subsection{Ablation studies}
 
\subsubsection{ACM and PLE effectiveness}
We perform ablation experiments on ACM to test the effectiveness of the intra- and inter-task relationships for both AGCN and AGCN++. 
As shown in Sec.~\ref{sec:acm}, $\mb{R}^t$ (Old-New) and $\mb{Q}^t$ (New-Old) are used to model inter-task label dependencies cross old and new tasks, $\mb{B}^t$ (New-New)  is used to model intra-task label dependencies, and while $\mb{R}^t$ and $\mb{Q}^t$ are unavailable, neither are $\mb{R}^{t-1}$ and $\mb{Q}^{t-1}$  in block $\mb{A}^{t-1}$, the $\mb{A}^{t-1}$ that inherit from the old task only build intra-task relationships.
As shown in Table~\ref{tab:ACM_ab}, if we do not build the label relationships across old and new tasks (w/o $\mb{R}^t$ \&  $\mb{Q}^t$), the performance of AGCN (Line 1 and 3) is already better than most non-AGCN methods.
For example, the comparison between AGCN (w/o $\mb{R}^t$ \& $\mb{Q}^t$) and LwF (w/o $\mb{A}^{t-1}$ \& $\mb{B}^t$ and $\mb{R}^t$  \& $\mb{Q}^t$) on mAP is: 38.05\% vs. 29.46\% (Split-WIDE, IL-MLCL), 49.47\% vs. 46.44\% (Split-WIDE, CL-MLCL), 31.52\% vs. 19.95\% (Split-COCO, IL-MLCL) and 44.53\% vs. 40.87\% (Split-COCO, CL-MLCL), as shown in Table~\ref{tab:ACM_ab}, Table~\ref{tab:results_WIDE} and Table~\ref{tab:results_COCO}. AGCN is also better than most non-AGCN methods on CF1 and OF1. 
This means only intra-task label relationships are effective for MLCL image recognition.
When the inter-task block matrices $\mb{R}^t$ and $\mb{Q}^t$ are available, AGCN with both intra- and inter-task relationships (Line 2 and 4) can perform even better in all three metrics.
For example, mAP comparisons of AGCN (w/ $\mb{A}^{t-1}$  \& $\mb{B}^t$, $\mb{R}^t$  \& $\mb{Q}^t$) and PRS on two datasets in two scenarios: 42.15\% vs. 39.70\% (Split-WIDE, IL-MLCL), 54.20\% vs. 51.42\% (Split-WIDE, CL-MLCL), 34.11\% vs. 31.08\% (Split-COCO, IL-MLCL) and 48.82\% vs. 46.39\% (Split-WIDE, CL-MLCL), as shown in Table~\ref{tab:ACM_ab}, Table~\ref{tab:results_WIDE} and Table~\ref{tab:results_COCO}, which means the inter-task relationships can enhance the multi-label recognition.

{And AGCN++ (w/ PLE) outperforms AGCN (w/o PLE) either (w/o $\mb{R}^t$ \& $\mb{Q}^t$) or (w/ $\mb{A}^{t-1}$  \& $\mb{B}^t$, $\mb{R}^t$  \& $\mb{Q}^t$) in all three metrics, which can prove the effectiveness of PLE. When w/o $\mb{R}^t$ \& $\mb{Q}^t$, for mAP, 42.25\% vs. 38.05\% (Split-WIDE, IL-MLCL), 54.72\% vs. 49.47\% (Split-WIDE, CL-MLCL), 35.19\% vs. 31.52\% (Split-COCO, IL-MLCL), 51.51\% vs. 44.53\% (Split-COCO, CL-MLCL). 
When w/ $\mb{A}^{t-1}$  \& $\mb{B}^t$, $\mb{R}^t$  \& $\mb{Q}^t$, for mAP, 45.73\% vs. 42.15\% (Split-WIDE, IL-MLCL), 57.07\% vs. 54.20\% (Split-WIDE, CL-MLCL), 38.23\% vs. 34.11\% (Split-COCO, IL-MLCL), 53.49\% vs. 48.82\% (Split-COCO, CL-MLCL).}


\subsubsection{Hyperparameter selection}
Then, we analyze the influences of loss weights and relationship-preserving loss on two benchmarks, as shown in Table~\ref{tab:weight}. 
When the relationship-preserving loss is unavailable, loss weight $\lambda_3$ is set to 0. The loss weights of others:  $\lambda_1=0.10$, $\lambda_2=0.90$ for Split-WIDE in IL-MLCL, $\lambda_1=0.70$, $\lambda_2=0.30$ for Split-WIDE in CL-MLCL, $\lambda_1=0.15$, $\lambda_2=0.85$ for Split-COCO in IL-MLCL and $\lambda_1=0.40$, $\lambda_2=0.60$ for Split-COCO in CL-MLCL, 
By adding the relationship-preserving loss $\ell_\text{gph}$, the performance gets more gains, and the values of forgetting are also lower, which means the mitigation of relationship-level catastrophic forgetting is quite essential for MLCL image recognition, and the relationship-preserving loss is effective. 
We select the best $\lambda_3$ as the hyper-parameters, \ie, $\lambda_3=10^4$ for Split-WIDE in IL-MLCL, $\lambda_3=10^3$ for Split-WIDE in CL-MLCL, $\lambda_3=10^4$ for Split-COCO in IL-MLCL and $\lambda_3=10^4$ for Split-COCO in CL-MLCL. 

\subsection{Visualization of ACM}
As shown in Fig.~\ref{fig:ACM_visual}, to verify the effectiveness of the constructed ACM, we offer the ACM visualizations on Split-WIDE and Split-COCO for IL-MLCL and CL-MLCL. 
We introduce the oracle augmented correlation matrix (oracle ACM) as the upper bound, which is constructed offline using hard label statistics of all tasks from corresponding datasets. \textbf{d} represents the Euclidean distance between the matrix and Oracle ACM. A smaller value of d means that the matrix is closer to oracle ACM, which proves that this ACM is better constructed.
As shown in Fig.~\ref{fig:ACM_visual}, the proposed ACM in both scenarios is close to the oracle ACM. This indicates constructing ACM with soft or hard label statistics is effective. {Note that in CL-MLCL, the ACM is constructed using only hard labels from the dataset, the ACMs of AGCN++ and AGCN are the same in the CL-MLCL scenario under the same dataset. And in IL-MLCL, the ACM is constructed with the soft labels produced by the model, so the ACM of AGCN++ is better constructed.} {In IL-MLCL, we can also observe that the ACM built in AGCN++ is closer to the Oracle ACM than in AGCN, 4.01\% vs. 4.18\% (Split-COCO) and 4.28\% vs. 4.54\% (Split-WIDE), which can prove that the PLE has reduced the accumulation of errors in the construction of label relationships.}

\section{Conclusion}

Multi-Label Continual Learning (MLCL) focuses on solving multi-label classification in continual learning.
It is challenging to construct convincing label relationships and reduce forgetting in MLCL because of the partial label problem.
This paper proposed a novel AGCN++ based on an auto-updated expert mechanism to solve the problem of  MLCL.
The key of our AGCN++ is to construct label relationships in a partial label data stream and reduce catastrophic forgetting to improve overall performance. 
We studied MLCL in both IL-MLCL and CL-MLCL scenarios.
In relationship construction, we showed the effectiveness of leveraging soft or hard label statistics to update the correlation matrix, even in the partial label data stream.
We also showed the effectiveness of {PLE in reducing the accumulation of errors in the construction of label relationships and suppressing forgetting.}
In terms of forgetting, we proposed an effective distillation loss and a novel relationship-preserving loss to mitigate class- and relationship-level forgetting.
Extensive experiments demonstrate that the proposed AGCN++ can capture well the label dependencies, thus achieving better MLCL performance in the IL-MLCL and CL-MLCL. 
Future work will study how to improve the construction of the old-old block using the correlation of only soft labels instead of inheriting the previously constructed ACM. It is believed that the performance will be further enhanced.

\section*{Acknowledgments}
This work was supported by the Natural Science Foundation of China (No. 61876121), the Postgraduate Research \& Practice Innovation Program of Jiangsu Province (No. SJCX21-1414), Suzhou Science and Technology Development Plan (Science and Technology Innovation for Social Development) Project (No. ss202133).

\ifCLASSOPTIONcaptionsoff
  \newpage
\fi



%


\bibliographystyle{IEEEtran}
\bibliography{citation.bib}

%




\end{document}